\documentclass[conference]{IEEEtran}
\usepackage{times}

\usepackage[numbers]{natbib}
\usepackage{multicol}
\usepackage[bookmarks=true]{hyperref}
\usepackage{amsmath}  
\usepackage{amssymb}
\usepackage{tabularx}
\usepackage{fancyhdr}
\usepackage{tikz}
\usetikzlibrary{shapes.geometric, arrows.meta, positioning, fit, backgrounds, calc, shadows}
\usepackage[utf8]{inputenc}

\usepackage{tcolorbox}
\usepackage{listings}
\tcbuselibrary{listings, skins, breakable}

\usepackage{booktabs}
\usepackage{graphicx}   
\usepackage[caption=false,font=footnotesize]{subfig}  
\usepackage[bookmarks=true]{hyperref}  

\newtcblisting{llmprompt}{
  colback=gray!5,
  colframe=gray!75!black,
  title=Prompt Template,
  listing only,
  listing options={
    basicstyle=\footnotesize\ttfamily,
    breaklines=true, 
    breakatwhitespace=false 
  },
  left=3mm,
  right=3mm,
  top=1mm,
  bottom=1mm,
  enhanced,
  breakable,
  width=\columnwidth
}

\newtcblisting{llmresponse}{
  colback=gray!5!blue!5,
  colframe=blue!75!black,
  title=LLM Response,
  listing only,
  listing options={
    basicstyle=\footnotesize\ttfamily,
    breaklines=true, 
    breakatwhitespace=false 
  },
  left=3mm,
  right=3mm,
  top=1mm,
  bottom=1mm,
  enhanced,
  breakable,
  width=\columnwidth
}

\usepackage[capitalise,noabbrev]{cleveref}
\usepackage{booktabs}   
\usepackage{multirow}   
\usepackage{array}      

\begin{document}

\title{Autonomous Reasoning for Spacecraft Control: A Large Language Model Framework with Group Relative Policy Optimization}


\author{\IEEEauthorblockN{Amit Jain}
\IEEEauthorblockA{\textit{Department of Aeronautics \& Astronautics} \\
\textit{Massachusetts Institute of Technology}\\
Cambridge, MA 02139 \\
amitjain@mit.edu}
\and
\IEEEauthorblockN{Richard Linares}
\IEEEauthorblockA{\textit{Department of Aeronautics \& Astronautics} \\
\textit{Massachusetts Institute of Technology}\\
Cambridge, MA 02139 \\
linaresr@mit.edu}
}

\fancypagestyle{firstpage}{%
  \fancyhf{}
  \renewcommand{\headrulewidth}{0pt}
  \fancyfoot[C]{\footnotesize 979-8-3315-7360-7/26/\$31.00 ©2026 IEEE}
}

\maketitle
\thispagestyle{firstpage}

\begin{abstract}
This paper presents a learning-based guidance-and-control approach that couples a reasoning-enabled Large Language Model (LLM) with Group Relative Policy Optimization (GRPO). A two-stage procedure consisting of Supervised Fine-Tuning (SFT) to learn formatting and control primitives, followed by GRPO for interaction-driven policy improvement, trains controllers for each environment. The framework is demonstrated on four control problems spanning a gradient of dynamical complexity, from canonical linear systems through nonlinear oscillatory dynamics to three-dimensional spacecraft attitude control with gyroscopic coupling and thrust constraints. Results demonstrate that an LLM with explicit reasoning, optimized via GRPO, can synthesize feasible stabilizing policies under consistent training settings across both linear and nonlinear
systems. The two-stage training methodology enables models to generate control sequences while providing human-readable explanations of their decision-making process. This work establishes a foundation for applying GRPO-based reasoning to autonomous control systems, with potential applications in aerospace and other safety-critical domains.
\end{abstract}

\section{Introduction}
\label{sec:introduction}

Space exploration increasingly demands autonomous capabilities as missions venture further from Earth and grow more complex. Traditional spacecraft control relies on ground-based operations and pre-programmed sequences, limiting adaptability and introducing vulnerability from communication delays. Recent advances in artificial intelligence, particularly Large Language Models (LLMs), present promising avenues for enhancing spacecraft autonomy. These models have demonstrated strong capabilities in reasoning, pattern recognition, and policy generation across diverse domains \citep{kim2024survey}.

Recent breakthrough work by \citet{zucchelli2025finetuned} demonstrated that fine-tuned LLMs can effectively control space systems while requiring less training data than traditional neural networks. Researchers have shown that fine-tuned LLMs like Llama-2 can generate high-precision thrust commands for spacecraft control across orbital transfers and lunar landings, while others demonstrated promising results in simulated environments like Kerbal Space Program \citep{carrasco2024finetuning}. Transformer-based reinforcement learning has also shown promise for multi-phase spacecraft trajectory optimization \citep{jain2025multi}. Prior work has demonstrated successful application of deep reinforcement learning to spacecraft docking maneuvers \citep{oestreich2021autonomous} and adaptive guidance with meta-learning \citep{gaudet2020adaptive}. However, these approaches typically operate in single-domain environments without cross-complexity transfer and treat control more as pattern recognition rather than reasoning-driven policy generation.

Despite these promising developments in LLM-based control, current research suffers from two critical limitations. First, there exists a lack of multi-environment generalization frameworks that can transfer control policies from simple academic systems to complex spacecraft dynamics while preserving reasoning capabilities. No systematic methodology exists for progressively scaling from fundamental control problems to aerospace applications while maintaining learned control policies and their associated reasoning processes. Second, while Group Relative Policy Optimization (GRPO) has shown remarkable success in mathematical reasoning and code generation tasks \citep{deepseek2024math, deepseek2025r1}, its application to control theory remains virtually unexplored, a significant missed opportunity for sample-efficient autonomous control.

Crucially, existing aerospace control systems lack the fundamental capability of \textit{autonomous reasoning}: the ability to analyze complex scenarios, provide decision justification, adapt to off-nominal conditions, and enable hierarchical decision-making. This represents a critical gap in current literature, as no prior work has developed reasoning-enabled models specifically for control applications. Unlike reactive approaches that rely on pre-programmed responses, reasoning-based controllers can dynamically assess system states, evaluate multiple control strategies, and provide transparent explanations for their selected actions, capabilities essential for next-generation aerospace systems operating in uncertain environments. Recent work has demonstrated that iterative reasoning approaches can enable efficient optimal control synthesis \citep{jain2025tiny}.

Group Relative Policy Optimization addresses limitations of traditional reinforcement learning by eliminating separate value function estimation while incentivizing structured reasoning \citep{aiengineering2024grpo, unsloth2025r1}. Rather than training a critic network, GRPO uses group-based normalization to compute advantages by comparing each sample against its peers, enabling sample-efficient policy improvement through relative performance comparisons. This approach naturally suits LLM reasoning architectures and significantly reduces memory requirements, a critical advantage for resource-constrained aerospace applications.

This paper introduces a reasoning-enabled framework that systematically applies GRPO to autonomous control problems. This approach represents a paradigm shift from reactive control to reasoning-driven policy generation, where controllers explicitly analyze system dynamics, evaluate control alternatives, and provide step-by-step justifications for their decisions. Foundational control knowledge is established through Supervised Fine-Tuning (SFT), where LLMs learn to reason about dynamical constraints before generating control policies. During this phase, models generate structured reasoning traces that justify their control decisions. GRPO then enables sample-efficient policy refinement through group-based comparisons, allowing reasoning principles to guide control generation.

The framework is demonstrated across four control problems that systematically increase in dynamical complexity, progressing from linear point-mass dynamics through nonlinear oscillatory systems to three-dimensional spacecraft operations involving continuous thrust optimization and multi-axis attitude stabilization. This progression validates the framework's ability to scale from fundamental control theory benchmarks to realistic aerospace applications while maintaining reasoning fidelity and control performance.

The remainder of this paper is organized as follows: Section \ref{sec:problem} formulates the problem of control using LLMs and GRPO; Section \ref{sec:methodology} details the methodology, including model architecture, training process, and implementation considerations; Section \ref{sec:results} presents simulation results across the four control scenarios; Section \ref{sec:conclusion} concludes with a summary of contributions and directions for future research.

\section{Problem Formulation}
\label{sec:problem}
To formalize the autonomous control challenge, control is cast as a sequential decision-making problem where an LLM-based agent must generate precise numerical outputs while satisfying physical constraints and providing interpretable explanations.

\subsection{Control as a Sequential Decision-Making Process}
\label{subsec:mdp}

The control of dynamical systems is formulated as a Markov Decision Process (MDP) defined by the tuple $(\mathcal{S}, \mathcal{A}, P, R, \gamma)$ \citep{sutton2018reinforcement}. The state space $\mathcal{S}$ is continuous, ranging from two-dimensional for the double integrator and Van der Pol oscillator ($\mathcal{S} \subset \mathbb{R}^2$) to three-dimensional for orbit raising and spacecraft detumbling ($\mathcal{S} \subset \mathbb{R}^3$). The action space $\mathcal{A}$ comprises control inputs constrained by actuator limits: scalar control for the simpler systems ($\mathcal{A} \subset \mathbb{R}$), thrust angle for orbital maneuvers ($\mathcal{A} \subset [0, 2\pi]$), and three-axis torque vectors for attitude control ($\mathcal{A} \subset \mathbb{R}^3$).

The state transition function $P: \mathcal{S} \times \mathcal{A} \times \mathcal{S} \to [0, 1]$ reduces to $s_{t+1} = f(s_t, a_t)$ for the deterministic systems considered here, where $f$ represents the governing differential equations. The reward function $R: \mathcal{S} \times \mathcal{A} \to \mathbb{R}$ quantifies controller performance based on state errors and control effort, while the discount factor $\gamma \in [0, 1]$ balances immediate versus future rewards. The control objective is to find a policy $\pi: \mathcal{S} \to \mathcal{A}$ that drives the system from arbitrary initial conditions to a target state while minimizing cost and satisfying constraints.

\subsection{System Dynamics Overview}
\label{subsec:dynamics}

The selection of evaluation systems follows a deliberate progression designed to validate the framework across increasingly challenging control regimes. The progression begins with systems that have well-understood analytical solutions, enabling direct comparison with optimal baselines, then advances to problems where nonlinearity, coupling, and multi-dimensional control introduce complexity representative of real aerospace applications.

The approach is evaluated on four representative dynamical systems:

\textbf{The double integrator} serves as the foundational benchmark, a fundamental linear system representing translational motion:
\begin{equation}
\ddot{x} = u
\end{equation}
State: $s = [x, \dot{x}]^T$. This system exhibits straightforward dynamics but requires careful control to avoid overshoot.

Moving beyond linear dynamics, \textbf{the Van der Pol oscillator} introduces nonlinear limit cycle behavior:
\begin{equation}
\ddot{x} - \mu(1-x^2)\dot{x} + x = u
\end{equation}
State: $s = [x, \dot{x}]^T$. The nonlinear damping term creates self-sustaining oscillations when uncontrolled, making stabilization
challenging.

Transitioning to aerospace-specific challenges, \textbf{the orbit raising problem} presents a spacecraft orbital transfer scenario with nonlinear gravitational and thrust dynamics:
\begin{equation}
\begin{aligned}
\dot{r} &= u \\
\dot{u} &= \frac{v^2}{r} - \frac{\mu}{r^2} + \frac{T\sin(\phi)}{m(t)} \\
\dot{v} &= -\frac{uv}{r} + \frac{T\cos(\phi)}{m(t)}
\end{aligned}
\end{equation}
State: $s = [r, u, v]^T$ represents orbital radius, radial velocity, and tangential velocity. Control: $\phi \in [0, 2\pi]$ is the thrust
angle. The time-varying mass $m(t) = m_0 + m_1 t$ (due to fuel consumption) and coupled gravitational-centrifugal dynamics create a challenging optimal control problem requiring precise thrust vectoring to achieve circular orbit insertion at increased altitude \citep{vallado2001fundamentals, miller2019low}.

\textbf{Spacecraft detumbling} represents the culmination of the complexity progression, combining nonlinear dynamics with multi-dimensional control through three-axis rotational dynamics:
\begin{equation}
J\dot{\omega} = -\omega \times (J\omega) + u
\end{equation}
where $\omega = [\omega_1, \omega_2, \omega_3]^T \in \mathbb{R}^3$ represents angular velocities, $J = \text{diag}([J_1, J_2, J_3])$ is the
inertia matrix, and $u = [u_1, u_2, u_3]^T \in \mathbb{R}^3$ is the control torque vector. The cross-product term introduces strong nonlinear coupling between axes \citep{wie2008space}.

These systems span a range of control challenges: linear versus nonlinear dynamics, scalar versus multi-dimensional control, decoupled versus strongly coupled state variables, and time-invariant versus time-varying parameters. The progression from double integrator through orbit
raising to detumbling systematically increases problem complexity while maintaining physical realism. Detailed system parameters, constraints, and bounds are provided in Section~\ref{sec:results}.
  
\subsection{LLM-Based Control Paradigm}
\label{subsec:llm_formulation}

The control problem is reformulated as a conditional text generation task suitable for Large Language Models. The key insight is that control problems can be naturally expressed in structured natural language that includes physical context, constraints, and objectives, information that LLMs can process through their pre-trained understanding of physics and mathematics.

Given the current system state $s_t \in \mathcal{S}$, the LLM receives a textual prompt describing the control scenario and generates a structured response containing two components:
\begin{enumerate}
\item \textbf{Reasoning:} Natural language explanation analyzing the control problem, system dynamics, and justification for the chosen strategy
\item \textbf{Control sequence:} Numerical control actions $\{a_t, a_{t+1}, \ldots, a_{t+N-1}\}$ formatted as structured text
\end{enumerate}

This formulation enables the policy $\pi_\theta$ to be parameterized by an LLM with parameters $\theta$, where:
\begin{equation}
\pi_\theta: \text{State Description} \to (\text{Reasoning}, \text{Control Sequence})
\end{equation}

The textual state description provides comprehensive context including system type, current state values with units, target state, time constraints, and control bounds. The specific encoding scheme and prompt templates are detailed in Section~\ref{subsec:state_action_repr}.

This paradigm offers several compelling advantages for autonomous control. The natural language reasoning enables human understanding and verification of control decisions, a capability critical for safety-critical aerospace applications where operators must trust and audit autonomous behavior. The same LLM architecture can be fine-tuned for different control problems without redesigning system-specific controllers, providing transferability across problem domains. Furthermore, LLMs can leverage their pre-trained understanding of physics and mathematics to generate control strategies that inherently respect physical constraints, while the pre-trained representations may enable learning from fewer environment interactions compared to training neural network controllers from scratch.

However, this approach introduces unique challenges: ensuring precise numerical outputs from a text generation model, maintaining output format consistency during reinforcement learning, and preventing generation of physically infeasible control sequences. The two-stage training methodology addresses these challenges through supervised fine-tuning followed by GRPO optimization.

\subsection{GRPO-Based Reinforcement Learning Objective}
\label{subsec:grpo_objective}

Traditional reinforcement learning aims to find a policy $\pi_\theta(a|s)$ parameterized by $\theta$ that maximizes expected cumulative discounted reward:
\begin{equation}
\label{eq:rl_objective}
J(\theta) = \mathbb{E}_{\tau \sim \pi_\theta}\left[\sum_{t=0}^{T} \gamma^t R(s_t, a_t)\right]
\end{equation}
where $\tau = (s_0, a_0, s_1, a_1, \dots, s_T, a_T)$ represents a trajectory sampled by following policy $\pi_\theta$.

For LLM-based control, this work adopts Group Relative Policy Optimization (GRPO), which eliminates the need for a separate value function network, significantly reducing memory requirements when fine-tuning large language models. The GRPO objective optimizes the policy through group-based advantage estimation:

\begin{equation}
\label{eq:grpo_loss}
\begin{aligned}
L_{\text{GRPO}}(\theta)
= & -\frac{1}{M} \sum_{i=1}^{M} \frac{1}{N} \sum_{j=1}^{N}
\min\!\Bigl(r_{ij}(\theta) A_{ij},\;
\text{clip}\!\bigl(r_{ij}(\theta), 1-\epsilon, \\ &  1+\epsilon\bigr) A_{ij}\Bigr) 
+ c \cdot L_{\text{KL}}(\theta \mid \theta_{\text{old}})
\end{aligned}
\end{equation}

where:
\begin{itemize}
    \item $r_{ij}(\theta) = \frac{\pi_\theta(a_{ij}|s_i)}{\pi_{\theta_{\text{old}}}(a_{ij}|s_i)}$ is the importance sampling ratio between the current policy and the policy used to generate responses.
    \item $A_{ij} = R(s_i, a_{ij}) - \bar{R}_i$ is the group-relative advantage, where $\bar{R}_i = \frac{1}{N} \sum_{k=1}^{N} R(s_i, a_{ik})$ is the mean reward across $N$ candidate control sequences for state $s_i$.
    \item $\text{clip}(r_{ij}(\theta), 1-\epsilon, 1+\epsilon)$ constrains the policy ratio to prevent destructively large updates (typically $\epsilon = 0.2$).
    \item $L_{\text{KL}}(\theta | \theta_{\text{old}})$ is a KL divergence penalty that regularizes the policy update (coefficient $c = 0.01$).
\end{itemize}

The key innovation of GRPO is the group-relative advantage $A_{ij}$, which compares each candidate control sequence against the mean performance of multiple candidates generated for the same state. This eliminates the need for a separate value function approximator while providing stable learning signals. The negative sign indicates that minimizing this loss maximizes the policy gradient objective.

The clipping mechanism, inspired by Proximal Policy Optimization \citep{schulman2017proximal}, ensures training stability by preventing rapid policy changes, while the KL penalty maintains proximity to the previous policy. This formulation is particularly well-suited for LLM fine-tuning, where memory constraints make Actor-Critic methods with separate value networks impractical.

\subsection{Control Objective and Performance Metrics}
\label{subsec:objectives}

For each control problem, the objective is to drive the system from an arbitrary initial state $s_0$ to the target state (typically the origin) within a fixed time horizon $T$, while minimizing a quadratic cost function:

\begin{equation}
\label{eq:control_cost}
J = \|s_T\|_{Q_f}^2 + \sum_{t=0}^{T-1} \left(\|s_t\|_{Q}^2 + \|a_t\|_{R}^2\right)
\end{equation}

where $Q_f, Q \succeq 0$ are state cost matrices penalizing deviation from the target, and $R \succ 0$ is the control cost matrix penalizing excessive control effort. Additionally, the following constraints must be satisfied:

\begin{itemize}
\item \textbf{Control bounds:} $a_t \in \mathcal{A}$ for all $t$, enforcing actuator limitations
\item \textbf{State constraints:} $s_t \in \mathcal{S}$ for all $t$, ensuring safe operation
\end{itemize}

Controller performance is evaluated through metrics that capture both terminal accuracy and trajectory quality. The final state error $\|s_T\|$ measures how precisely the controller achieves the target state, while the total trajectory cost $J$ from Eq.~\ref{eq:control_cost} quantifies the cumulative penalty incurred during the maneuver. Constraint satisfaction rate indicates the percentage of trajectories completing without violations, essential for safety-critical applications. Control effort $\sum_{t=0}^{T-1} \|a_t\|^2$ quantifies actuator usage, relevant for fuel-limited spacecraft operations, and convergence quality assesses the rate of state error reduction throughout the trajectory, distinguishing between controllers that converge smoothly versus those exhibiting oscillatory behavior.

These metrics enable comprehensive evaluation of the LLM-based controller against optimal control baselines \citep{jain2023sparse}, assessing both control performance and constraint compliance. The specific cost matrix values and constraint bounds for each system are provided in Section~\ref{sec:results}.

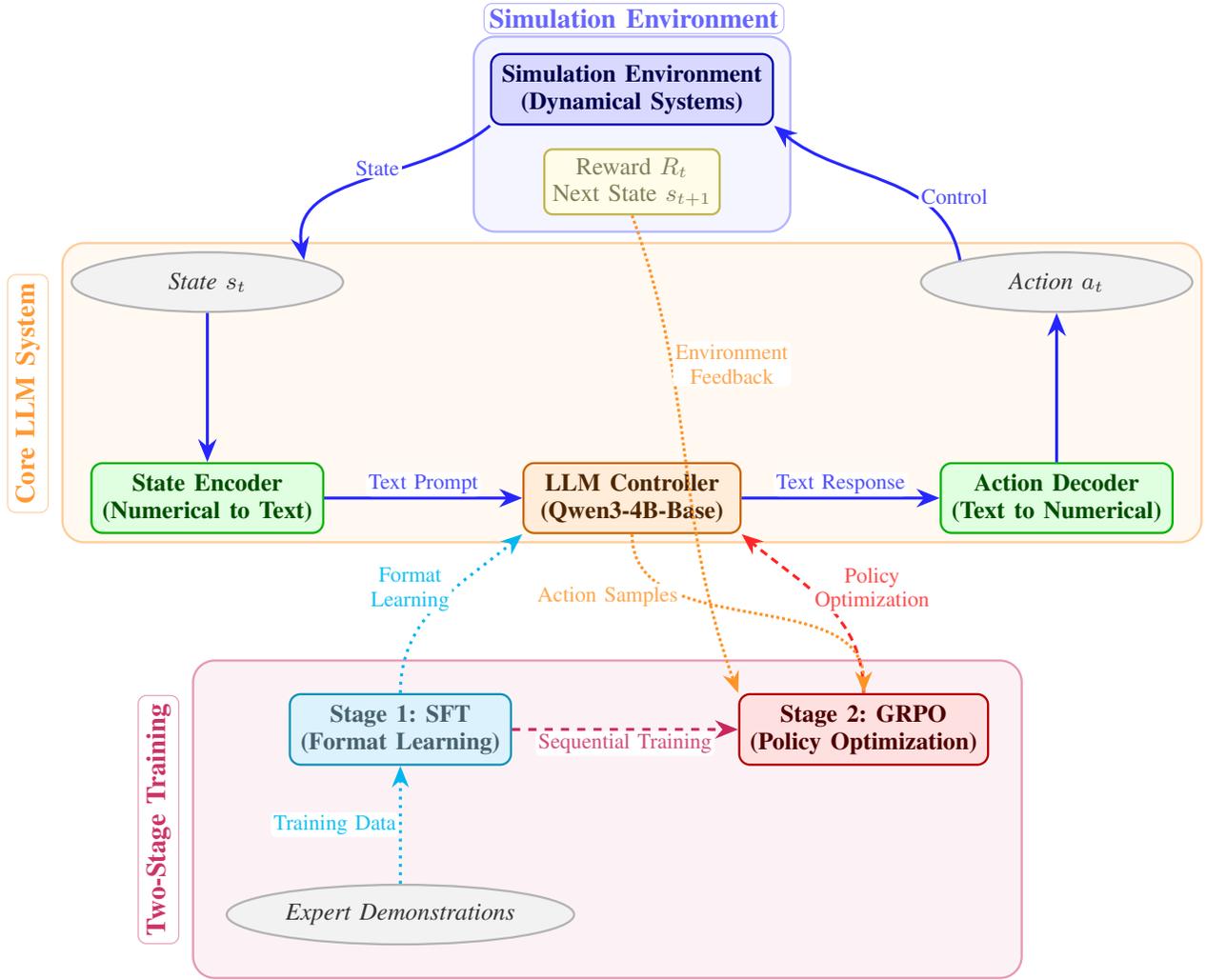
\begin{figure*}[htb!]
  \centering
  \resizebox{0.92\textwidth}{!}{%
  \begin{tikzpicture}[
      node distance=0.6cm and 1.2cm, 
      auto,
      block/.style={
          rectangle, draw=black!60, thick, rounded corners=4pt,
          minimum height=2.4em, minimum width=6.5em,
          text centered, font=\small\bfseries, align=center,
          inner sep=4pt
      },
      simenv/.style={
          block, fill=blue!15, draw=blue!70!black, thick, 
          minimum width=7.5em, minimum height=2.4em,
          text=blue!30!black
      },
      llm/.style={
          block, fill=orange!15, draw=orange!80!black, thick, 
          minimum width=8em, minimum height=2.4em,
          text=orange!30!black
      },
      process/.style={
          block, fill=green!12, draw=green!70!black, thick, 
          minimum width=7em, minimum height=2.4em,
          text=green!30!black
      },
      io/.style={
          ellipse, draw=gray!60, thick, fill=gray!10,
          minimum height=2.2em, minimum width=10em,
          text centered, font=\small\itshape,
          text=gray!40!black, inner sep=3pt
      },
      sft/.style={
          block, fill=cyan!12, draw=cyan!70!black, thick, 
          minimum width=6.5em, minimum height=2.2em,
          text=cyan!30!black
      },
      grpo/.style={
          block, fill=red!12, draw=red!70!black, thick, 
          minimum width=6.5em, minimum height=2.2em,
          text=red!30!black
      },
      feedback/.style={
           rectangle, draw=yellow!70!black, thick, fill=yellow!10, rounded corners=3pt,
           minimum height=2em, minimum width=6em, text centered, 
           font=\small, align=center, text=yellow!40!black, inner sep=3pt
      },
      main_arrow/.style={-{Stealth[length=3mm, width=2.4mm]}, line width=1.1pt, draw=blue!85},
      training_arrow/.style={-{Stealth[length=3mm, width=2.4mm]}, line width=1.1pt, draw=purple!85, dashed},
      sft_arrow/.style={-{Stealth[length=2.8mm, width=2.2mm]}, line width=1.1pt, draw=cyan!85, dotted},
      grpo_arrow/.style={-{Stealth[length=2.8mm, width=2.2mm]}, line width=1.1pt, draw=red!85, densely dashed},
      feedback_arrow/.style={-{Stealth[length=2.8mm, width=2.2mm]}, line width=1.1pt, draw=orange!85, densely dotted}
  ]

  \node [simenv] (sim) at (0,7.5) {Simulation Environment \\ (Dynamical Systems)};
  \node [io] (state_in) at (-5.5,5) {State $s_t$};
  \node [feedback] (reward_info) at (0,6.3) {Reward $R_t$ \\ Next State $s_{t+1}$};
  \node [io] (action_out) at (5.5,5) {Action $a_t$};
  \node [process] (state_enc) at (-5.5,2.2) {State Encoder \\ (Numerical to Text)}; 
  \node [llm] (llm_ctrl) at (0,2.2) {LLM Controller \\ (Qwen3-4B-Base)}; 
  \node [process] (action_dec) at (5.5,2.2) {Action Decoder \\ (Text to Numerical)};
  \node [sft] (sft) at (-3,-0.8) {Stage 1: SFT \\ (Format Learning)};
  \node [grpo] (grpo) at (3,-0.8) {Stage 2: GRPO \\ (Policy Optimization)};
  \node [io] (expert_data) at (-3,-3.2) {Expert Demonstrations};

  \draw [main_arrow] (sim.south west) to[out=-140,in=100] 
        node[pos=0.35, left, font=\footnotesize, text=blue!70, fill=white, rounded corners=2pt, inner sep=1pt] {State} (state_in.north east);
  \draw [main_arrow] (state_in.south) -- (state_enc.north);
  \draw [main_arrow] (state_enc.east) -- 
        node[pos=0.5, above, font=\footnotesize, text=blue!70, fill=white, rounded corners=2pt, inner sep=1pt] {Text Prompt} (llm_ctrl.west);
  \draw [main_arrow] (llm_ctrl.east) -- 
        node[pos=0.5, above, font=\footnotesize, text=blue!70, fill=white, rounded corners=2pt, inner sep=1pt] {Text Response} (action_dec.west);
  \draw [main_arrow] (action_dec.north) -- (action_out.south);
  \draw [main_arrow] (action_out.north west) to[out=100,in=-40] 
        node[pos=0.35, right, font=\footnotesize, text=blue!70, fill=white, rounded corners=2pt, inner sep=1pt] {Control} (sim.south east);

  \draw [feedback_arrow] (reward_info.south) to[out=-70,in=110] 
        node[pos=0.3, right, font=\footnotesize, text=orange!70, align=center, fill=white, rounded corners=2pt, inner sep=1pt] {Environment \\ Feedback} (grpo.north west);

  \draw [sft_arrow] (expert_data.north) -- 
        node[pos=0.5, left, font=\footnotesize, text=cyan!70, fill=white, rounded corners=2pt, inner sep=1pt] {Training Data} (sft.south);
  \draw [sft_arrow] (sft.north) to[out=90,in=-140] 
        node[pos=0.6, left, font=\footnotesize, text=cyan!70, align=center, fill=white, rounded corners=2pt, inner sep=1pt] {Format \\ Learning} (llm_ctrl.south west);
  \draw [grpo_arrow] (grpo.north) to[out=90,in=-40] 
        node[pos=0.6, right, font=\footnotesize, text=red!70, align=center, fill=white, rounded corners=2pt, inner sep=1pt] {Policy \\ Optimization} (llm_ctrl.south east);
  
  \draw [training_arrow] (sft.east) -- 
        node[pos=0.5, below, font=\footnotesize, text=purple!70, fill=white, rounded corners=2pt, inner sep=1pt] {Sequential Training} (grpo.west);

  \draw [feedback_arrow] (llm_ctrl.south) to[out=-90,in=90] 
        node[pos=0.3, left, font=\footnotesize, text=orange!70, fill=white, rounded corners=2pt, inner sep=1pt] {Action Samples} (grpo.north);

  \begin{pgfonlayer}{background}
    \node [fit=(sim)(reward_info), 
           draw=blue!30, thick, fill=blue!4, rounded corners=8pt, 
           inner sep=6pt] (fit_sim) {};
    \node[font=\bfseries, anchor=north west, xshift=4pt, yshift=12pt,
          fill=white, draw=blue!30, rounded corners=3pt, inner sep=2pt, 
          text=blue!60] 
         at (fit_sim.north west) {Simulation Environment};
    
    \node [fit=(state_enc)(llm_ctrl)(action_dec)(state_in)(action_out), 
           draw=orange!40, thick, fill=orange!5, rounded corners=8pt, 
           inner sep=3pt] (fit_core) {};
    \node[font=\bfseries, anchor=north west, xshift=-20pt, yshift=-100pt, 
          fill=white, draw=orange!40, rounded corners=3pt, inner sep=3pt,
          text=orange!80, rotate=90] 
         at (fit_core.north west) {Core LLM System};
    
    \node [fit=(sft)(grpo)(expert_data), 
           draw=purple!40, thick, fill=purple!5, rounded corners=8pt, 
           inner sep=12pt] (fit_training) {};
    \node[font=\bfseries, anchor=north west, xshift=-20pt, yshift=-105pt,
          fill=white, draw=purple!40, rounded corners=3pt, inner sep=3pt,
          text=purple!80, rotate=90] 
         at (fit_training.north west) {Two-Stage Training};
  \end{pgfonlayer}

  \end{tikzpicture}%
  } 
  \caption{Enhanced system architecture for LLM-based control with two-stage training. The \textcolor{blue!60}{\textbf{simulation environment}} provides dynamic system context and feedback. The \textcolor{orange!60}{\textbf{core LLM system}} handles runtime inference through state encoding, LLM processing, and action decoding. The \textcolor{purple!60}{\textbf{two-stage training system}} implements sequential supervised fine-tuning (SFT) for format learning, followed by Group Relative Policy Optimization (GRPO) for policy refinement.}
  \label{fig:system_architecture}
\end{figure*}

\section{Methodology}
\label{sec:methodology}
Having established the theoretical framework for LLM-based control with GRPO optimization, this section details its practical implementation. The system architecture, model selection, training procedures, and reward function design that translate the formulation of Section~\ref{sec:problem} into a working control system are described below.

\subsection{System Architecture}
\label{subsec:architecture}

The proposed system integrates an LLM with a two-stage training approach for autonomous control of dynamical systems, as illustrated in Figure~\ref{fig:system_architecture}. At the core sits an LLM Controller that receives system state as textual input and generates control action sequences with accompanying reasoning. Before reaching the LLM, a State Encoder converts numerical state vectors into formatted text prompts that preserve physical meaning. After generation, an Action Decoder extracts numerical control sequences from the LLM's text responses using regular expressions with robust error handling for malformed outputs.

The controller interacts with Simulation Environments that model each dynamical system, providing state transitions through numerical integration and computing rewards based on trajectory quality. Training proceeds in two stages: supervised fine-tuning first establishes proper output formatting and basic control intuition from expert demonstrations, then GRPO reinforcement learning refines the policy through environment interaction. This architecture enables end-to-end learning from state observations to control outputs while maintaining interpretability through natural language reasoning.

\subsection{Base Model Selection and Parameter-Efficient Fine-Tuning}
\label{subsec:llm_selection}

Qwen3-4B-Base \citep{qwen2024} is selected as the foundation model due to its balance of 4 billion parameters and computational efficiency. Through the two-stage training approach combining supervised fine-tuning and GRPO, the model is trained to generate structured control outputs that include natural language explanations of control strategies. While the model produces reasoning-like explanations in the \texttt{<REASONING>} sections, these are learned through the training process rather than emergent reasoning capabilities.

\subsubsection{LoRA-Based Fine-Tuning}

Rather than fine-tuning all model parameters, this work employs Low-Rank Adaptation (LoRA)~\cite{hu2022lora}, a parameter-efficient fine-tuning technique that introduces trainable low-rank decomposition matrices into transformer layers while keeping pre-trained weights frozen: $W' = W_0 + \Delta W = W_0 + BA$, where $W_0 \in \mathbb{R}^{d \times k}$ represents the frozen pre-trained weights, and $B \in \mathbb{R}^{d \times r}$, $A \in \mathbb{R}^{r \times k}$ are trainable low-rank matrices with rank $r \ll \min(d,k)$.

This parameter-efficient approach offers critical advantages for control applications. LoRA reduces trainable parameters by over 99\% (from 4B to ~40M parameters with rank $r=8$), enabling fine-tuning on standard GPU hardware with only 12GB GPU memory compared to 80GB+ for full fine-tuning. Training convergence is significantly faster, with SFT completing in 2-3 hours per system compared to days for full fine-tuning. LoRA adapters (typically 100-200MB) can be saved separately from the base model, allowing multiple specialized control adapters to be efficiently switched for different systems. By freezing base model weights, LoRA maintains the LLM's general reasoning capabilities while adding task-specific control knowledge, preventing catastrophic forgetting.

All experiments use LoRA with rank $r = 8$ applied to query and value projection layers in all attention blocks, with LoRA alpha parameter set to 16 and dropout rate of 0.05. This configuration results in fewer than 1\% of base model parameters being trainable, dramatically reducing memory requirements while maintaining capacity to learn task-specific control policies. The relatively low rank is sufficient for structured control tasks, as the adaptation primarily involves learning numerical output patterns and constraint handling rather than acquiring fundamentally new linguistic capabilities.

\subsection{State and Action Representation}
\label{subsec:state_action_repr}

\subsubsection{State Encoding Scheme}

Numerical state vectors $s_t \in \mathcal{S}$ are converted into structured natural language prompts that provide comprehensive context for control decision-making. Each prompt contains five key components: system specification identifying the control problem type, current state information presenting numerical values with appropriate units, target state specification defining the desired final state, temporal constraints specifying the time horizon and number of control steps, and control specifications enumerating actuator bounds and state constraints. This structured encoding ensures the LLM receives comprehensive context while maintaining a standardized format that facilitates learning across diverse control tasks. Importantly, the encoding preserves physical meaning by using descriptive terms such as ``angular velocity'' rather than abstract labels, allowing the model to leverage its pre-trained understanding of physical concepts.



\subsubsection{Action Decoding Scheme}

The LLM generates responses with two main sections: \texttt{<REASONING>} containing natural language explanation of the control strategy, and \texttt{<CONTROLS>} containing the numerical control sequence formatted as comma-separated values for scalar control or bracketed vectors for multi-dimensional control. Robust parsing with regular expressions extracts numerical values and automatically enforces constraints by clipping values to actuator bounds, verifying correct sequence length, and handling parsing failures through fallback mechanisms that apply safe default controls during inference or reject samples with negative reward penalties during training.










\subsection{Prompt and Response Format}
\label{subsec:prompt_format}

The LLM receives control problems in a structured natural language format and generates responses containing both reasoning and numerical control sequences. To illustrate the format, consider a spacecraft attitude detumbling scenario:

\textbf{System Prompt:}
\begin{llmprompt}
You are a spacecraft control systems expert.

Given a spacecraft detumbling maneuver with initial angular velocities [omega_1, omega_2, omega_3], generate a sequence of 10 3D torque vectors to bring the spacecraft to rest (omega = [0,0,0]) in exactly 5.00 seconds.

DYNAMICS: omega_dot = -J^(-1)(omega x J*omega) + J^(-1)*u     with J = diag([J1, J2, J3])

CONSTRAINTS: |omega_i| <= omega_max rad/s, 
            |u_i| <= u_max N*m

Explain your approach between <REASONING> and </REASONING>.
Then provide exactly 10 torque vectors as comma-separated values between <CONTROLS> and </CONTROLS>.
Format each torque vector as [u1, u2, u3] with 3 decimal places.
\end{llmprompt}

\textbf{User Prompt (Problem Instance):}
\begin{llmprompt}
Control a spacecraft detumbling maneuver with initial angular velocities [omega_1=0.350, omega_2=-0.520, omega_3=0.180] rad/s to bring the spacecraft to rest (omega = [0,0,0]) in 5.00 seconds using 10 steps. Inertia matrix: J = diag([14.0, 10.0, 8.0]) kg*m^2.
Keep angular velocities within [-1, 1] rad/s and torques within [-4, 4] N*m.
\end{llmprompt}

\textbf{Expected LLM Response:}
\begin{llmresponse}
<REASONING>
For this spacecraft detumbling maneuver starting with angular velocities [omega_1=0.350, omega_2=-0.520, omega_3=0.180] rad/s, I'm using BVP optimal control to bring the spacecraft to rest in 5.00 seconds.

The spacecraft dynamics follow Euler's rotational equations:
omega_dot = -J^(-1)(omega x J*omega) + J^(-1)*u
with inertia matrix J = diag([14.0, 10.0, 8.0]) kg*m^2.

Analysis:
- Initial angular momentum magnitude: 0.649 rad/s
- Dominant tumbling axis: Y (omega_2)
- Coupling constants: K_1=-0.143, K_2=0.600, K_3=-0.250

Strategy: optimal
- Apply 3D torque sequence over 10 steps
- Each step duration: 0.5s
- Target: Zero angular velocity (detumbled state)
- Constraints: |omega_i| <= 1 rad/s, |u_i| <= 4.0 N*m

This approach exploits the nonlinear coupling between axes while minimizing control effort and respecting physical constraints.
</REASONING>

<CONTROLS>
[-1.245, 2.187, -0.658]
[-1.089, 1.923, -0.542]
[-0.876, 1.534, -0.398]
[-0.654, 1.142, -0.267]
[-0.445, 0.782, -0.154]
[-0.267, 0.478, -0.068]
[-0.134, 0.245, -0.012]
[-0.045, 0.089, 0.021]
[-0.008, 0.019, 0.012]
[0.000, 0.001, 0.002]
</CONTROLS>
\end{llmresponse}

This structured format enables the LLM to generate both interpretable reasoning about control strategy and precise numerical control sequences. The \texttt{<REASONING>} section provides transparency into the model's decision-making process, while the \texttt{<CONTROLS>} section contains the actionable control inputs for execution.

\subsection{Expert Demonstration Generation}
\label{subsec:data_generation}

High-quality expert demonstrations form the foundation for supervised fine-tuning. Each control system utilizes an appropriate optimal control method \citep{kirk2004optimal}: LQR for linear systems (double integrator), sequential quadratic programming for nonlinear systems (Van der Pol), and boundary value problem solvers based on Pontryagin's Maximum Principle for rotational dynamics (spacecraft).

The data generation process samples 2000 initial states per system uniformly within state space bounds ($s_0 \sim \mathcal{U}(s_{\min}, s_{\max})$), split into 1800 training and 200 evaluation samples. System-specific solvers compute optimal control sequences minimizing the quadratic cost with matrices $Q = I$, $R = 0.1 \cdot I$, $Q_f = 10 \cdot I$. To enhance generalization, the dataset incorporates diverse strategies: optimal solutions (40\%), alternative strategies prioritizing energy efficiency or time optimality (30\%), suboptimal but practical approaches (20\%), and recovery scenarios for challenging situations (10\%). Each control sequence is simulated through true system dynamics using appropriate numerical integration (Euler for linear systems, RK45 for Van der Pol's stiff nonlinear dynamics) to verify constraint satisfaction, and trajectories are annotated with performance metrics including cost, final state error, control effort, smoothness, and constraint violations.

Rather than using fixed templates, reasoning text is generated dynamically by analyzing specific initial state characteristics including distance from target, velocity direction, system-specific challenges, strategy justifications with explicit trade-offs, and predicted trajectory behavior. For example, high-velocity states receive reasoning emphasizing aggressive braking to prevent overshoot, while low-energy states emphasize gentle convergence with minimal effort. This state-dependent approach ensures the model develops genuine understanding of control principles rather than memorizing fixed patterns. The final dataset contains 1800 training and 200 evaluation samples per system, with average reasoning length of 150-250 words and control sequences of 10 steps.

\subsection{Two-Stage Training Approach}
\label{subsec:two_stage_training}

The methodology employs a two-stage approach that separates format learning from control optimization, ensuring the model first learns proper output structure before engaging in policy refinement through environmental interaction.

\subsubsection{Stage 1: Supervised Fine-Tuning (SFT)}

The first stage establishes the model's understanding of control problem formatting using expert demonstrations. The SFT stage trains the LLM to minimize negative log-likelihood of expert demonstrations: $\mathcal{L}_{\text{SFT}} = -\frac{1}{N} \sum_{i=1}^{N} \log \pi_\theta(y_i | x_i)$, where $x_i$ is the input prompt and $y_i$ is the expert response. The objectives are to produce properly formatted responses with consistent \texttt{<REASONING>} and \texttt{<CONTROLS>} sections, develop basic control intuition from optimal examples, establish numerical precision for control outputs, and learn to parse natural language control specifications.

Training uses AdamW optimizer (8-bit) with learning rate $2 \times 10^{-4}$, linear warmup for 50 steps followed by cosine decay, batch size of 4 sequences, and 500 training steps over sequences up to 1024 tokens with gradient clipping at max norm 1.0. Models are evaluated on the held-out evaluation set to verify format compliance rate (target: $>$ 95\%), parsing success rate, basic control performance through LQR cost comparison, and constraint satisfaction on evaluation trajectories. Only models meeting format compliance and basic performance thresholds proceed to GRPO training.

\subsubsection{Stage 2: Group Relative Policy Optimization (GRPO)}

The second stage refines control policies through environmental interaction and reward-based learning, implementing the GRPO objective defined in Eq.~\ref{eq:grpo_loss}. The objectives are to optimize control performance through reward signals from trajectory simulations, maintain output format consistency established during SFT, discover strategies that may outperform initial demonstrations, and balance exploration with exploitation.

Training uses AdamW optimizer with learning rate $3 \times 10^{-6}$, cosine annealing with 100-step warmup, batch size of 1 state per step, generating $N=4$ candidate control sequences per state with temperature 1.0 and min-p filtering (0.1) over 500 training steps. The clipping parameter is $\epsilon = 0.2$ and KL penalty coefficient is $c = 0.01$. Each training step executes the following procedure: sample initial states from the training distribution, generate $N=4$ candidate control sequences using the current policy with temperature sampling, simulate each candidate through the system dynamics using RK45 integration, calculate trajectory rewards for each candidate, compute group mean reward $\bar{R}_i = \frac{1}{N}\sum_{j=1}^{N}R(s_i, a_{ij})$ and advantages $A_{ij} = R(s_i, a_{ij}) - \bar{R}_i$, compute importance sampling ratios $r_{ij}(\theta) = \frac{\pi_\theta(a_{ij}|s_i)}{\pi_{\theta_{\text{old}}}(a_{ij}|s_i)}$, calculate the GRPO loss using Eq.~\ref{eq:grpo_loss}, and update policy parameters using gradient descent while monitoring format compliance.

The group-based advantage estimation provides stable learning signals by comparing each candidate against the mean performance, eliminating the need for a separate value function. The clipping mechanism prevents destructive policy updates by constraining $r_{ij}(\theta)$ to $[1-\epsilon, 1+\epsilon]$.

\subsection{Reward Function Design}
\label{subsec:reward_design}

The reward function balances multiple objectives critical for effective control, computed as a weighted sum: $R(s, a) = w_1 R_{\text{LQR}} + w_2 R_{\text{terminal}} + w_3 R_{\text{constraint}} + w_4 R_{\text{format}} + w_5 R_{\text{auxiliary}}$. The primary performance metric is the LQR cost $R_{\text{LQR}} = -\sum_{t=0}^{T} (s_t^T Q s_t + u_t^T R u_t)$ with cost matrices $Q = I$, $R = 0.1 \cdot I$, $Q_f = 10 \cdot I$. Terminal accuracy is rewarded through exponential and stepped bonuses based on final state error $\|s_T\|$. Constraint satisfaction is enforced through violation penalties and validity bonuses. Format compliance rewards ensure output structure consistency with penalties for format violations. Auxiliary components include control persistence penalties that discourage excessive control near the target and convergence bonuses that reward monotonic error decrease.

The component weights follow a curriculum schedule: early GRPO training (steps 0-200) uses high format reward weight and lower control weight; mid training (steps 200-400) uses balanced weights; late training (steps 400-500) uses low format weight and high control weight. This gradually shifts focus from maintaining output structure to optimizing control performance while preventing format collapse. Specific weight values and reward ranges for each system are provided in Section~\ref{sec:results}.

\section{Experimental Results}
\label{sec:results}
With the training methodology established, this section evaluates the framework's effectiveness across the four control systems introduced in Section~\ref{subsec:dynamics}, progressing from linear dynamics through increasingly complex aerospace applications. For each system, the GRPO training progression is demonstrated, control performance is analyzed, and example reasoning outputs that illustrate the model's decision-making process are presented.

Tables~\ref{tab:experimental_params} and \ref{tab:system_params} summarize the experimental configuration. The framework employs Unsloth for memory-efficient LoRA fine-tuning with the TRL library for both SFT and GRPO implementations.

\begin{table}[ht]
\centering
\caption{Experimental Parameters}
\label{tab:experimental_params}
\begin{tabular}{@{}llr@{}}
\toprule
\textbf{Category} & \textbf{Parameter} & \textbf{Value} \\
\midrule
\multicolumn{3}{l}{\textit{Model Configuration}} \\
& Base Model & Qwen3-4B-Base \\
& Fine-tuning Method & LoRA (rank 8) \\
& Trainable Parameters & $\sim$40M ($<$ 1\%) \\
\midrule
\multicolumn{3}{l}{\textit{Dataset}} \\
& Training Samples/System & 1800 \\
& Evaluation Samples/System & 200 \\
\midrule
\multicolumn{3}{l}{\textit{SFT Training}} \\
& Learning Rate & $2 \times 10^{-4}$ \\
& Batch Size & 4 \\
\midrule
\multicolumn{3}{l}{\textit{GRPO Training}} \\
& Learning Rate & $1 \times 10^{-6}$ \\
& Batch Size & 1 \\
& Candidates per State & 8 \\
& Temperature & 1.0 \\
& Clipping ($\epsilon$) & 0.2 \\
& KL Coefficient ($c$) & 0.05 \\
\bottomrule
\end{tabular}
\end{table}

\begin{table}[ht]
\centering
\caption{System-Specific Parameters and Constraints}
\label{tab:system_params}
\begin{tabular}{@{}lll@{}}
\toprule
\textbf{System} & \textbf{Category} & \textbf{Values} \\
\midrule
Double  & State: $x, \dot{x}$ & $[-1, 1]$ \\
Integrator & Control: $u$ & $[-3, 3]$ \\
& Training: SFT/GRPO & 1000/1000 steps \\
\midrule
Van der Pol & State: $x, \dot{x}$ & $[-1, 1]$ \\
& Control: $u$ & $[-3, 3]$ \\
& Parameter: $\mu$ & $1.0$ \\
& Training: SFT/GRPO & 1000/500 steps \\
\midrule
Orbit Raising & State: $r, u, v$ & $[0.5, 2.0]$, $[-0.5, 0.5]$, $[0.5, 1.5]$ \\
& Control: $\phi$ & $[0, 2\pi]$ rad \\
& Parameters: $\mu, T$ & $1.0$, $0.1405$ N \\
& Mass: $m(t)$ & $m_0=1.0$, $m_1=-0.075$ kg/s \\
& Training: SFT/GRPO & 1000/1000 steps \\
\midrule
Spacecraft & State: $\omega_i$ $(i=1,2,3)$ & $[-1.0, 1.0]$ rad/s \\
Detumbling & Control: $u_i$ $(i=1,2,3)$ & $[-4, 4]$ N$\cdot$m \\
& Inertia: $J$ & diag$([14, 10, 8])$ kg$\cdot$m$^2$ \\
& Training: SFT/GRPO & 1000/3000 steps \\
\bottomrule
\end{tabular}
\end{table}


\begin{figure*}[htb!]
  \centering
  \subfloat[GRPO training progress\label{fig:di_grpo}]{%
    \includegraphics[width=0.35\textwidth]{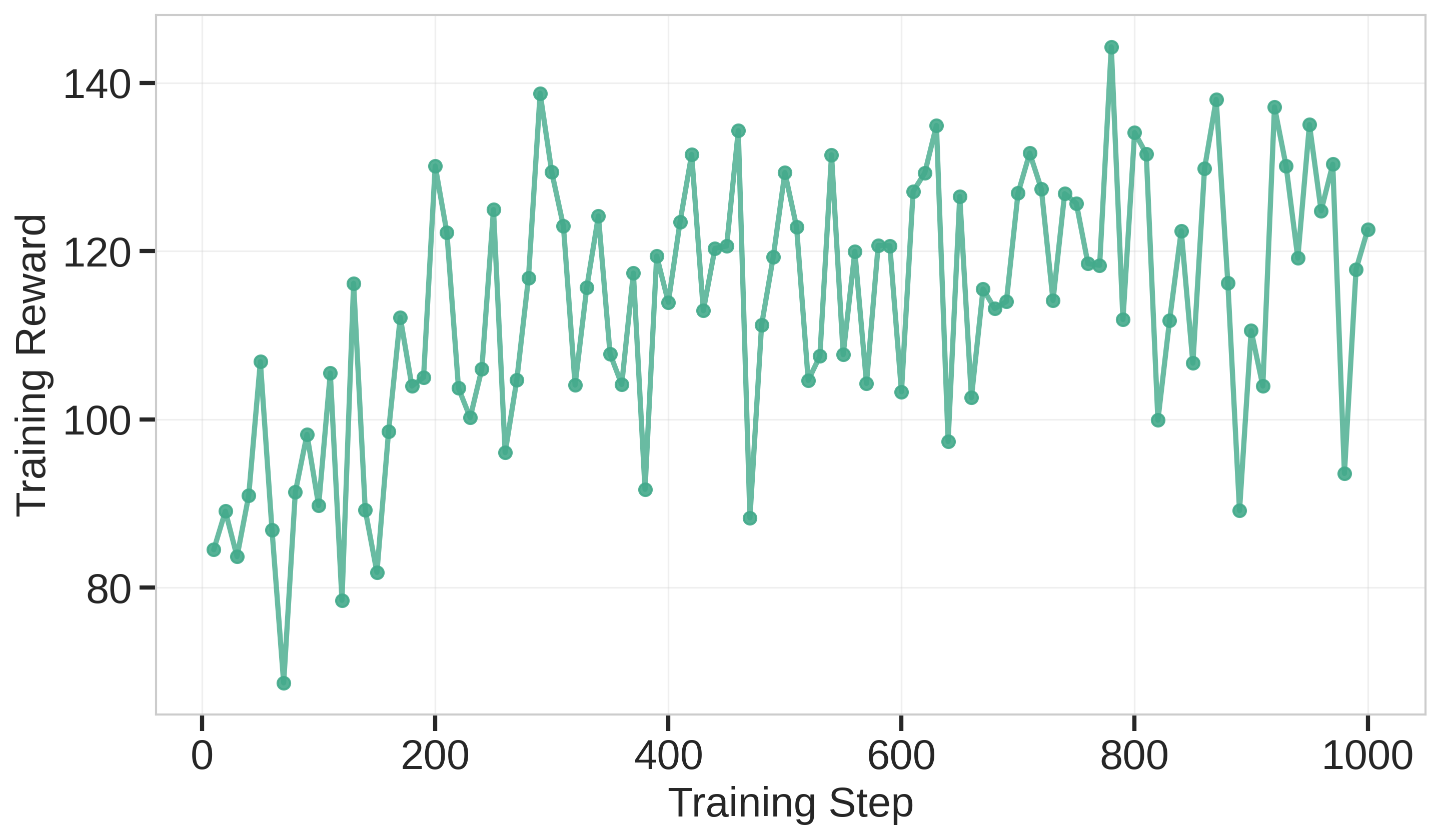}%
  }\hspace{1cm}%
  \subfloat[Control trajectories\label{fig:di_control}]{%
    \includegraphics[width=0.35\textwidth]{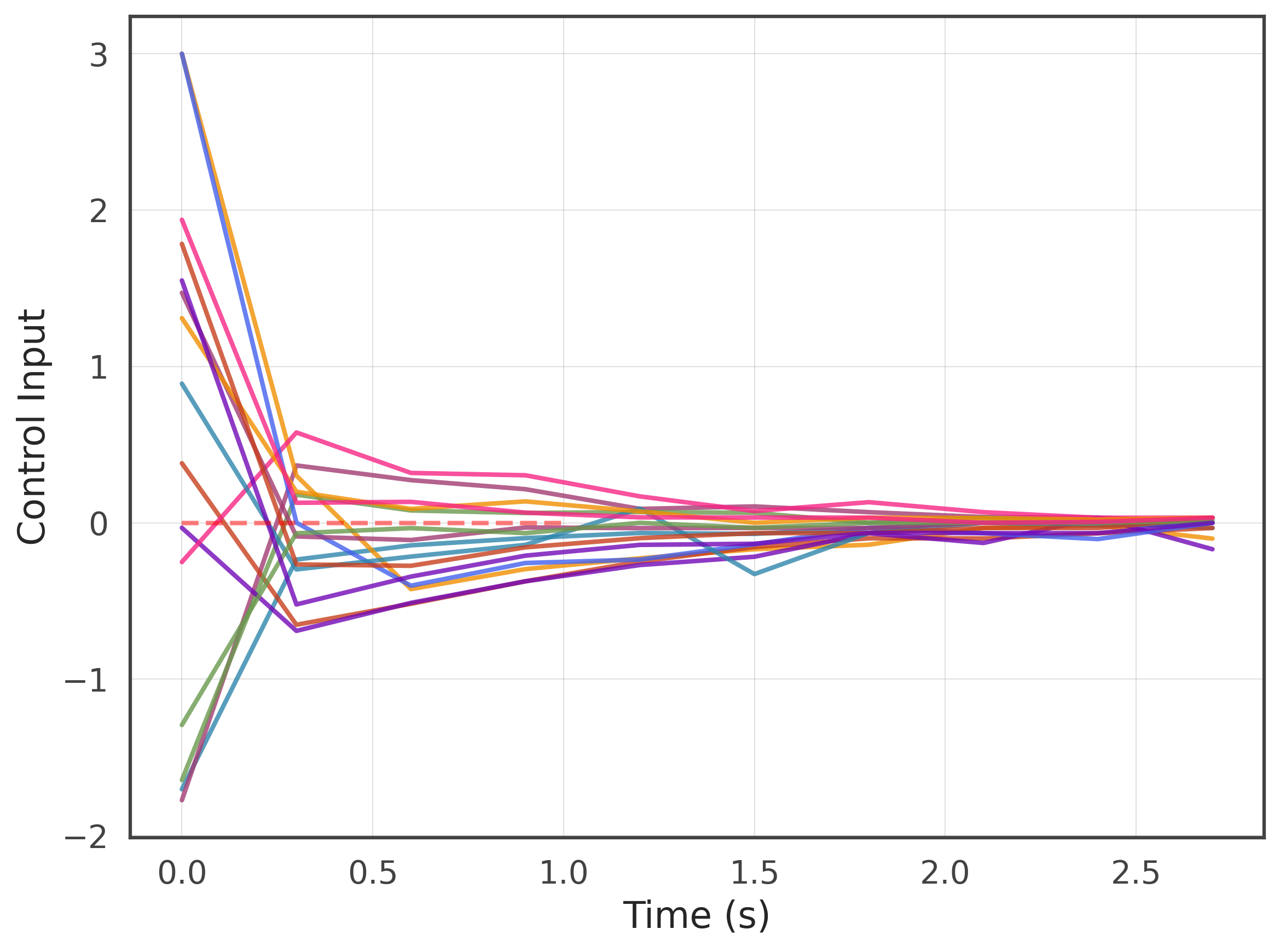}%
  }

  \vspace{0.2cm}

  \subfloat[Position trajectories\label{fig:di_position}]{%
    \includegraphics[width=0.35\textwidth]{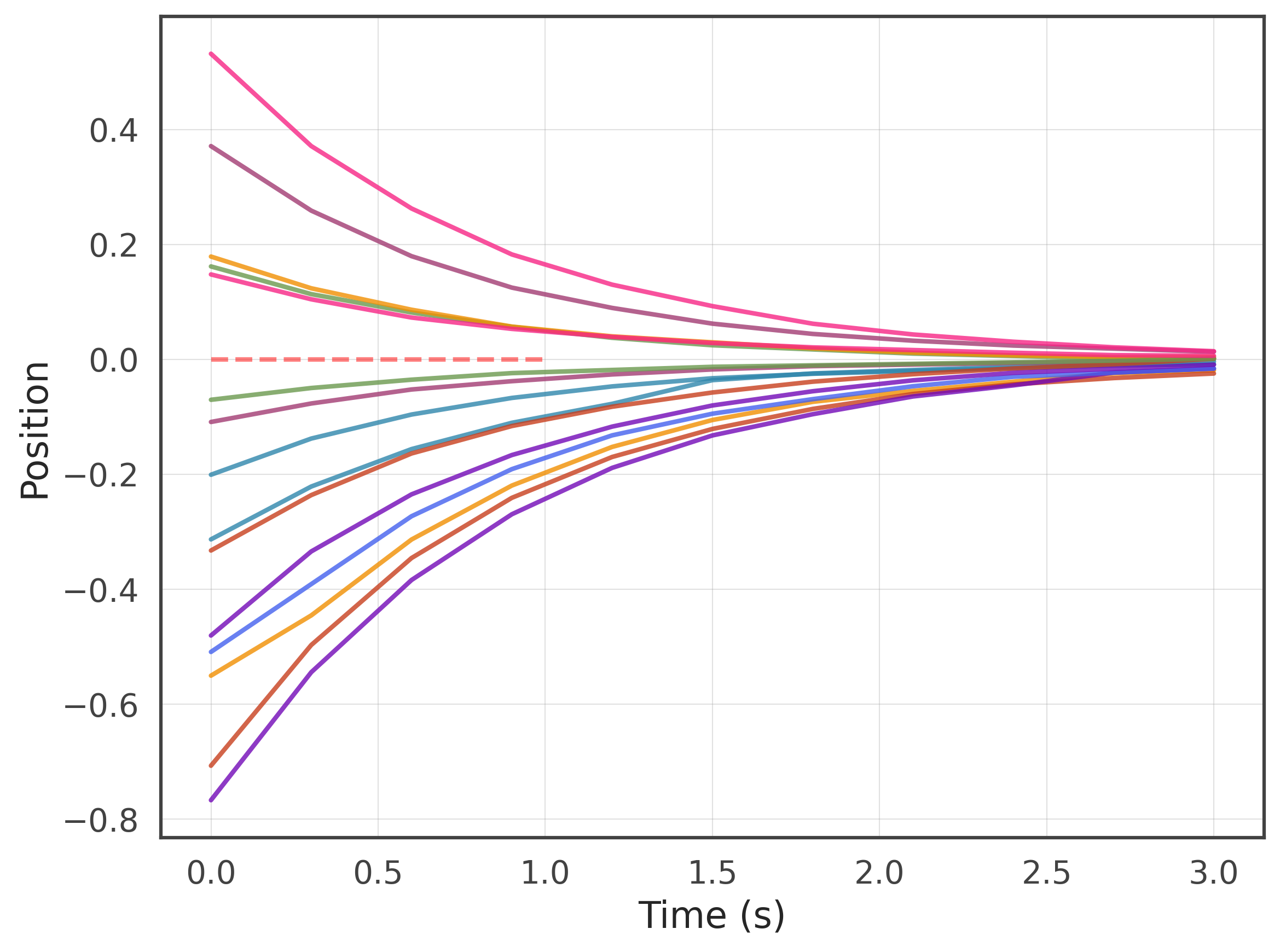}%
  }\hspace{1cm}%
  \subfloat[Velocity trajectories\label{fig:di_velocity}]{%
    \includegraphics[width=0.35\textwidth]{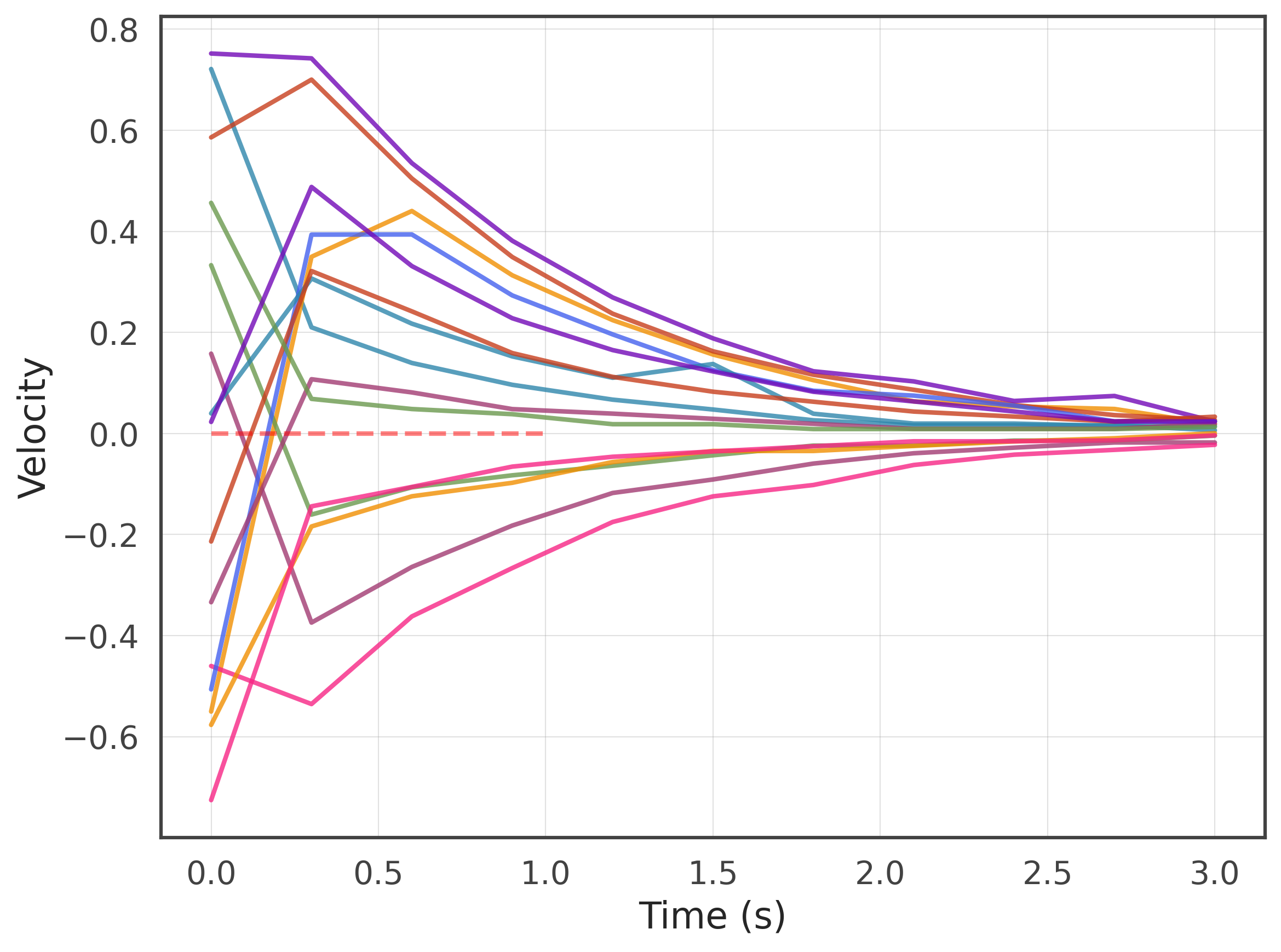}%
  }

  \caption{Double integrator control performance: (a) GRPO training progression showing reward convergence, (b) control input trajectories, (c) position state evolution, and (d) velocity state evolution.}
  \label{fig:di_results}
\end{figure*}

\subsection{Double Integrator System}

The evaluation begins with the double integrator, which serves as a canonical linear control problem to validate the fundamental capabilities of the LLM-GRPO framework before advancing to more challenging nonlinear systems. Figure~\ref{fig:di_results} presents the training progression and control performance. Figure~\ref{fig:di_grpo} demonstrates GRPO training convergence over 1000 steps, with rewards exhibiting a clear upward trend from near-zero to approximately 10-12 after 200 steps, maintaining sustained performance thereafter. The variance decreases significantly after the initial 100 steps, indicating stable policy improvement despite minor fluctuations from the stochastic optimization process.

The control and state trajectories in Figures~\ref{fig:di_control}-\ref{fig:di_velocity} demonstrate effective stabilization across multiple evaluation episodes. Control inputs exhibit initial transients with magnitudes up to $\pm 2$ N, followed by modulated oscillatory behavior during the first 2 seconds before settling near zero, all remaining within the $[-3, 3]$ N bounds. Position and velocity states converge to the origin from diverse initial conditions ($[-0.4, 0.3]$ for position, $[-0.4, 0.4]$ m/s for velocity) within 3-4 seconds, exhibiting monotonic or near-monotonic well-damped behavior characteristic of optimal LQR-type policies.

\begin{figure*}[htb!]
  \centering
  \subfloat[GRPO training progress\label{fig:vo_grpo}]{%
    \includegraphics[width=0.35\textwidth]{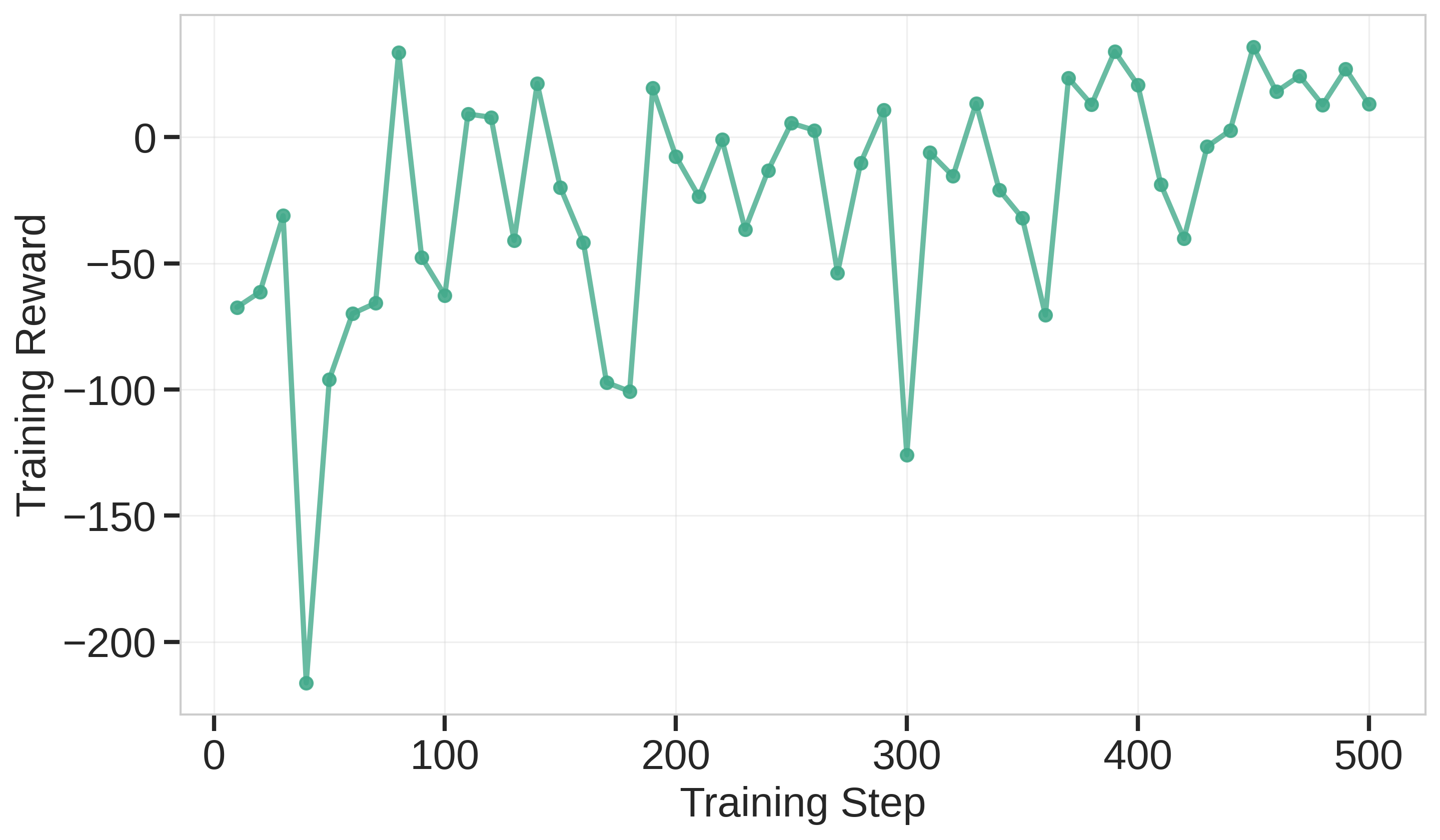}%
  }\hspace{1cm}%
  \subfloat[Control trajectories\label{fig:vo_control}]{%
    \includegraphics[width=0.35\textwidth]{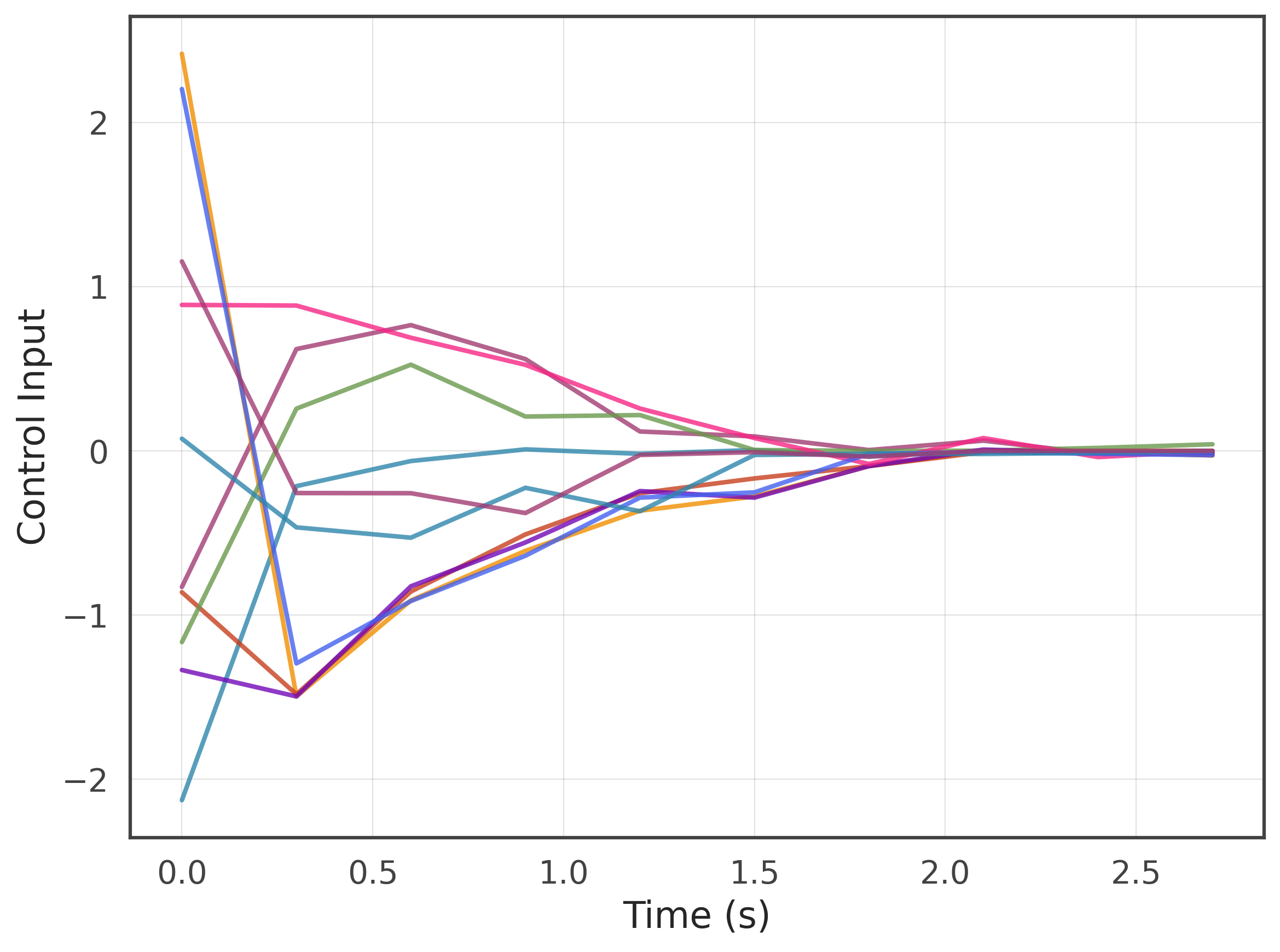}%
  }

  \vspace{0.2cm}

  \subfloat[Position trajectories\label{fig:vo_position}]{%
    \includegraphics[width=0.35\textwidth]{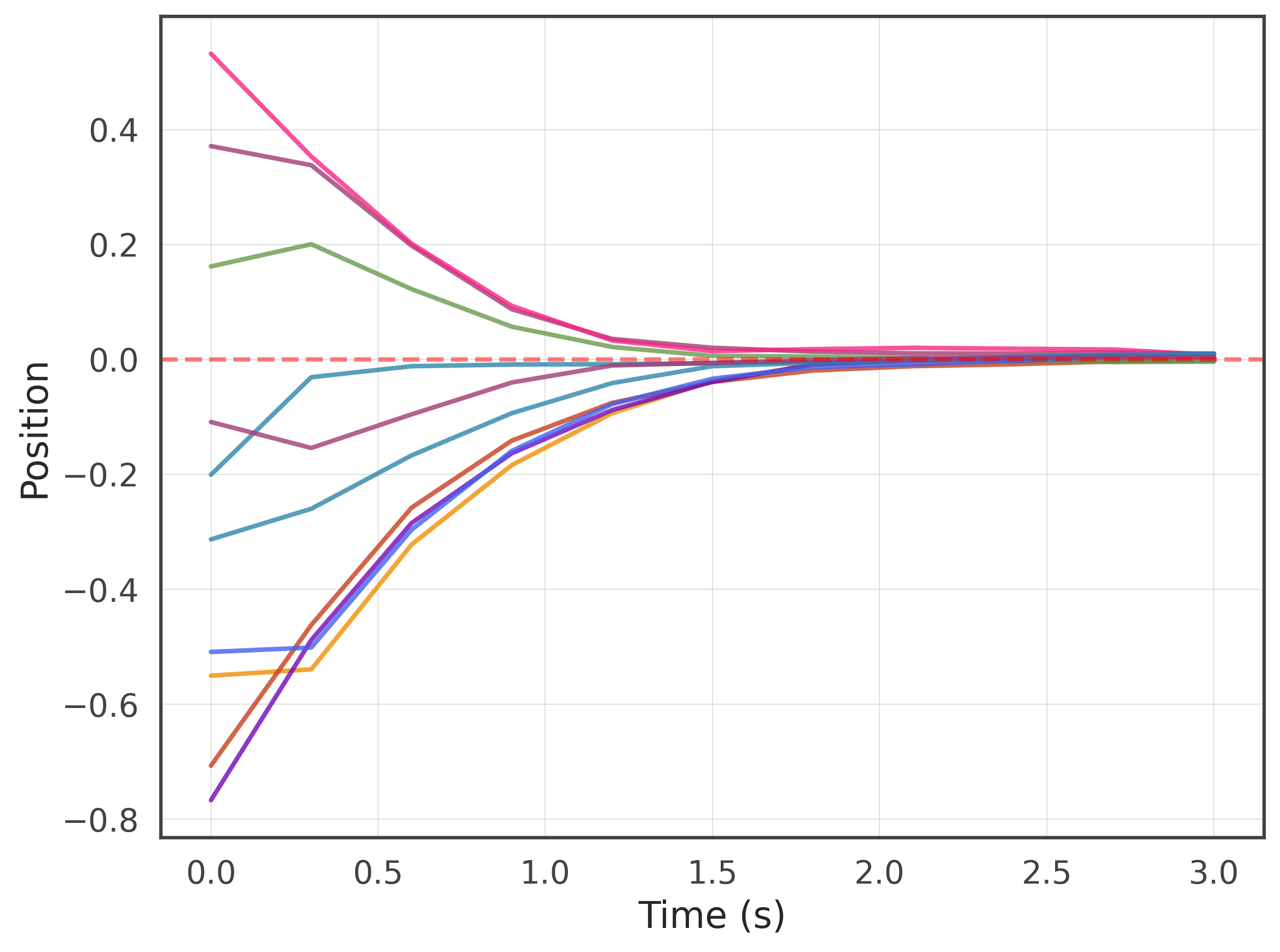}%
  }\hspace{1cm}%
  \subfloat[Velocity trajectories\label{fig:vo_velocity}]{%
    \includegraphics[width=0.35\textwidth]{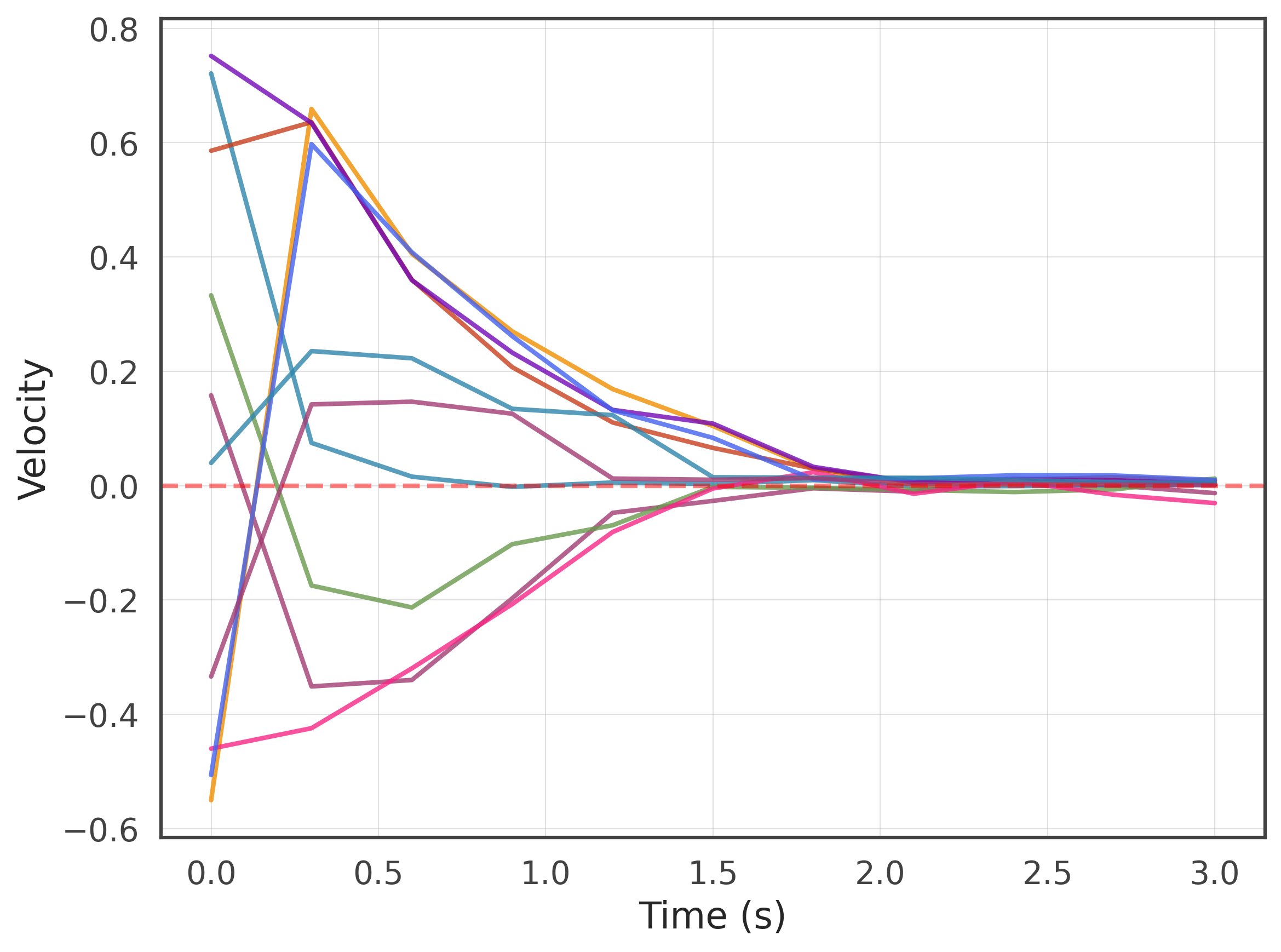}%
  }

  \caption{Van der Pol oscillator control performance: (a) GRPO training progression with nonlinear dynamics, (b) control input trajectories for limit cycle suppression, (c) position state evolution, and (d) velocity state evolution.}
  \label{fig:vo_results}
\end{figure*}

\subsection{Van der Pol Oscillator}

Having validated the framework on linear dynamics, the analysis now examines whether it can handle nonlinear systems. The Van der Pol oscillator introduces limit-cycle dynamics with $\mu = 1.0$, where the system naturally sustains oscillations that require active damping for stabilization. Figure~\ref{fig:vo_results} presents the training and control performance. Figure~\ref{fig:vo_grpo} shows training rewards starting near $-50$ with a dramatic improvement around step 200, reaching near-zero levels and gradually refining to slightly positive values by step 600. The training exhibits higher variance than the double integrator, with occasional dips below $-100$ during the first 400 steps, reflecting the additional complexity of nonlinear dynamics and limit-cycle suppression.

The control and state trajectories in Figures~\ref{fig:vo_control}-\ref{fig:vo_velocity} reveal more complex behavior than the linear case. Control inputs span nearly $[-2.5, 2.5]$ N with non-monotonic oscillatory profiles, maintaining sustained control effort beyond 2 seconds before settling near zero, all within the $[-3, 3]$ N constraints. Position and velocity states from diverse initial conditions converge within 4-6 seconds, exhibiting transient oscillations and velocity peaks up to $\pm 0.6$ m/s characteristic of limit-cycle suppression. The variability in convergence paths highlights the state-dependent nature of the nonlinear dynamics.

\subsection{Orbit Raising Maneuver}

\begin{figure}[htb!]
  \centering
  \includegraphics[width=0.9\columnwidth]{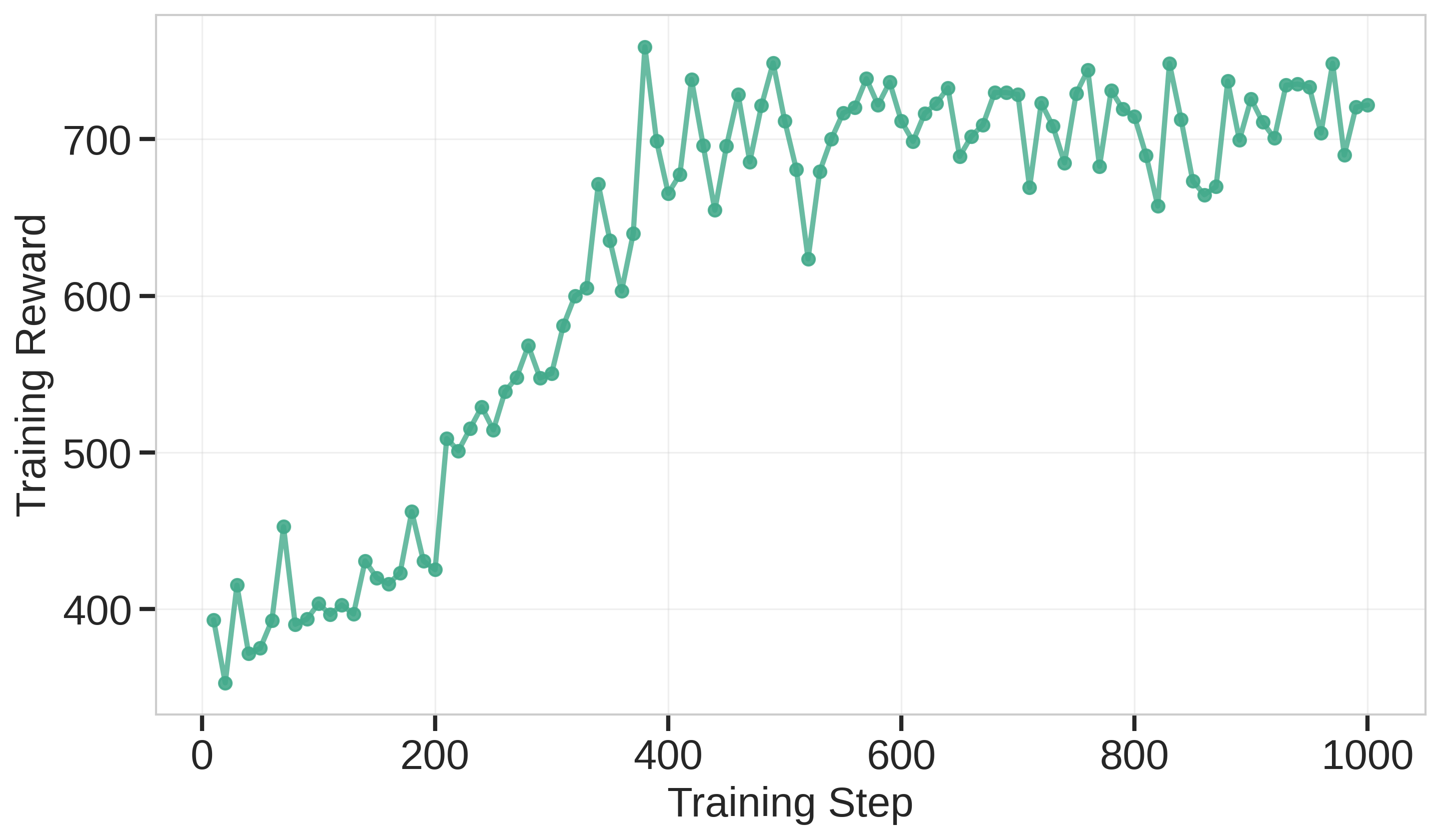}
  \caption{Orbit raising GRPO training progression showing reward convergence for orbital transfer control.}
  \label{fig:orbit_grpo}
\end{figure}

\begin{figure*}[htb!]
  \centering
  \subfloat[Radial distance trajectories\label{fig:orbit_radius}]{%
    \includegraphics[width=0.35\textwidth]{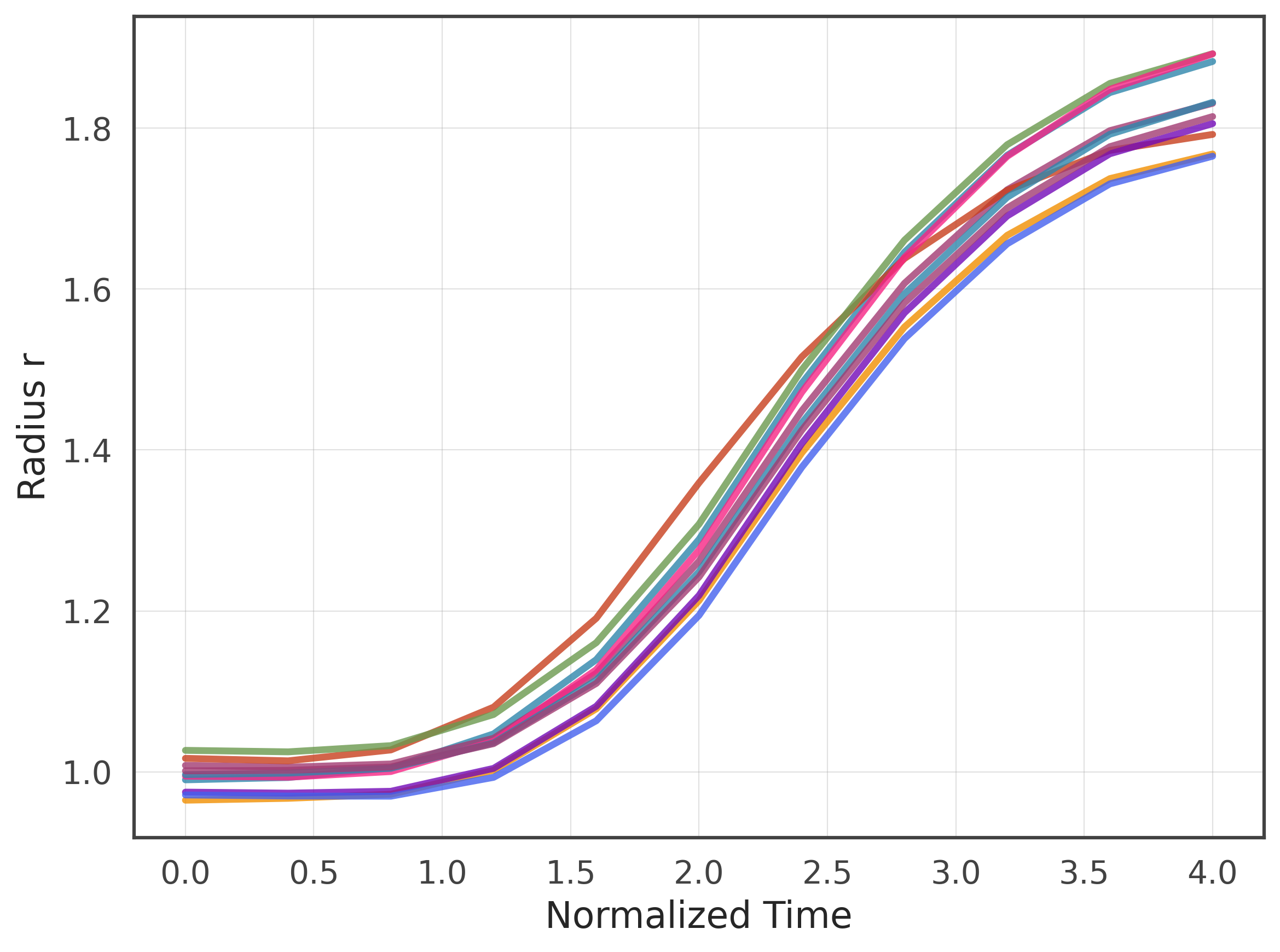}%
  }\hspace{1cm}%
  \subfloat[Radial velocity trajectories\label{fig:orbit_radial_vel}]{%
    \includegraphics[width=0.35\textwidth]{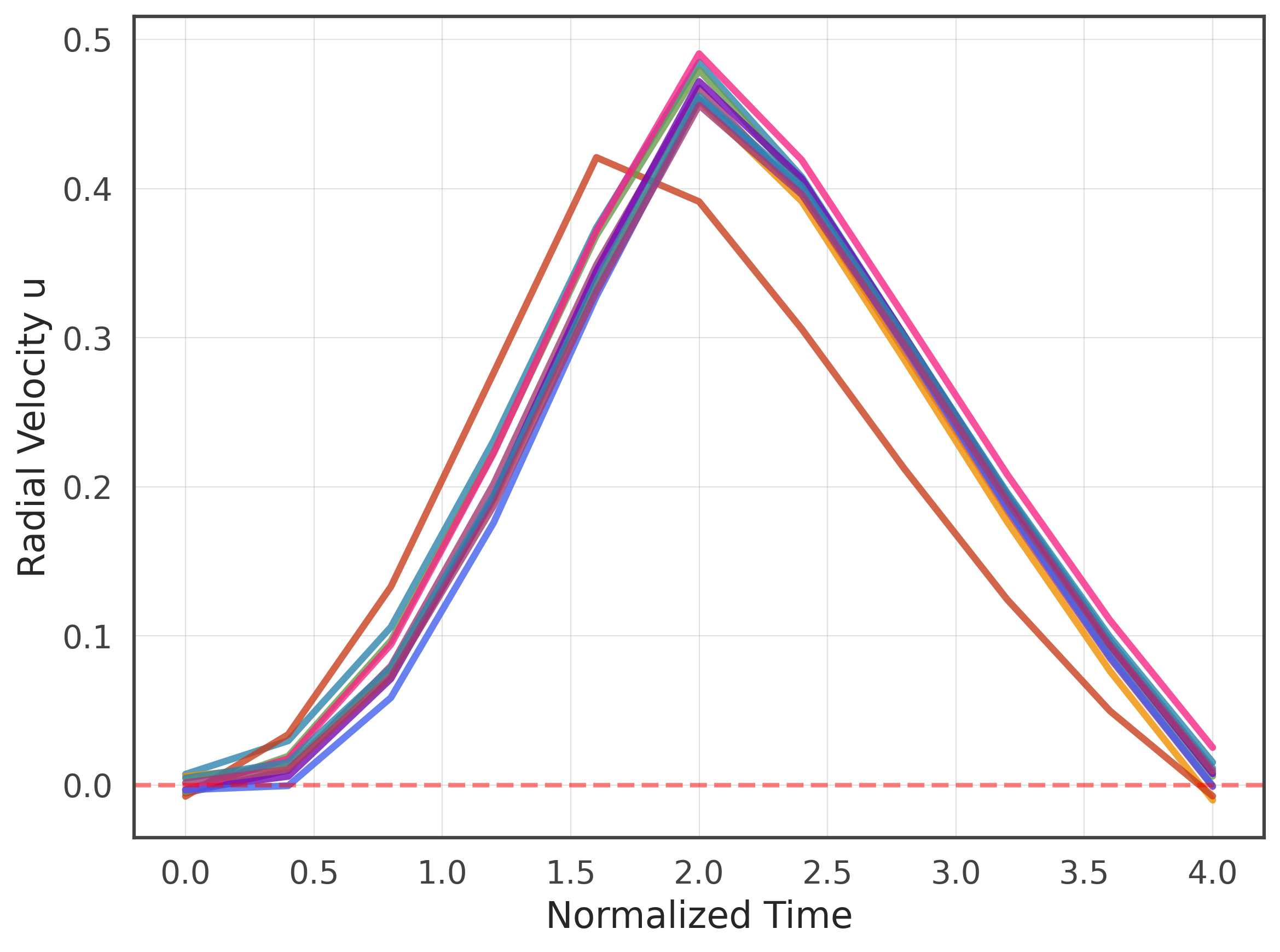}%
  }

  \vspace{0.2cm}

  \subfloat[Tangential velocity trajectories\label{fig:orbit_tangential_vel}]{%
    \includegraphics[width=0.35\textwidth]{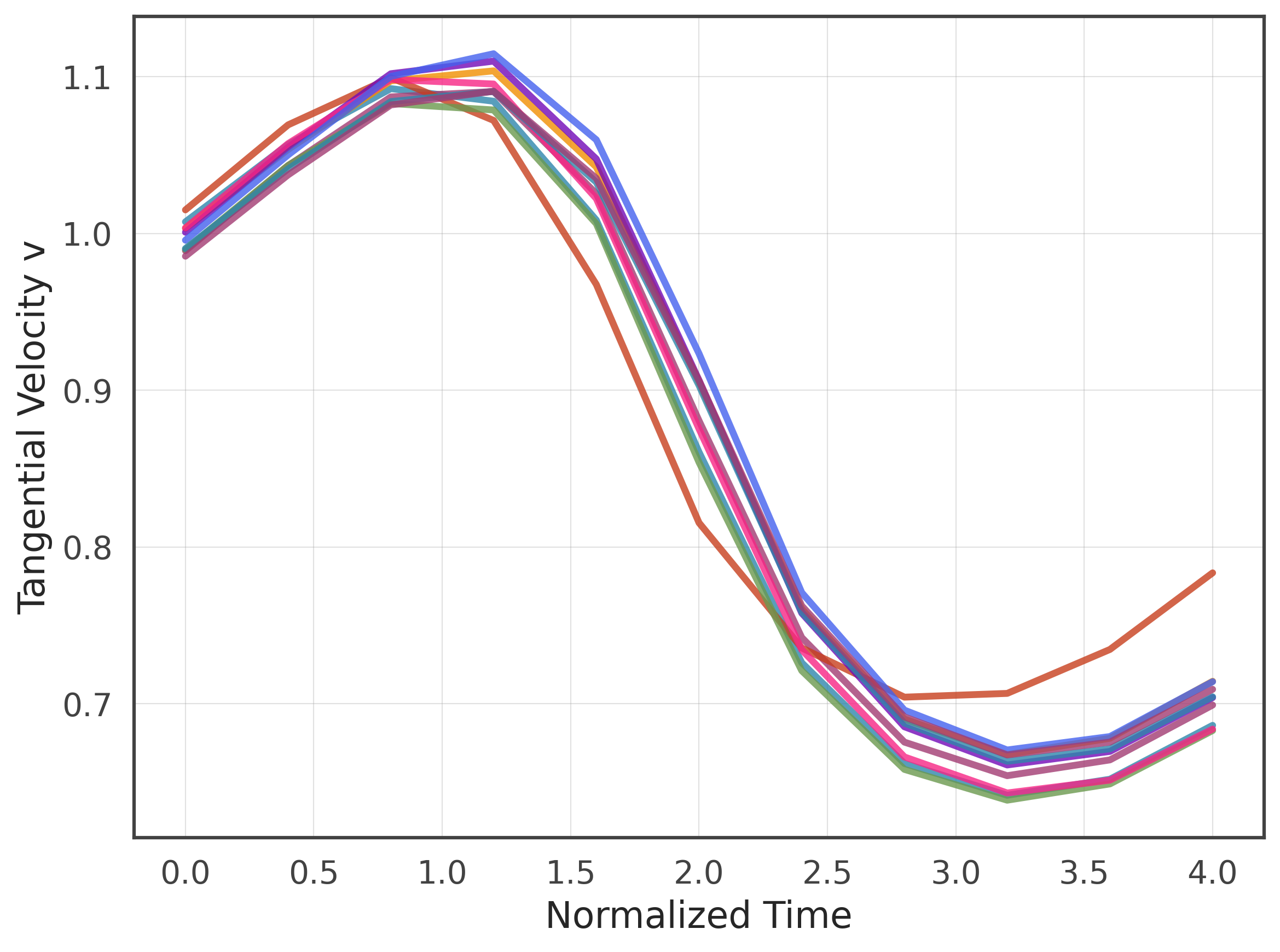}%
  }\hspace{1cm}%
  \subfloat[Thrust angle trajectories\label{fig:orbit_thrust}]{%
    \includegraphics[width=0.35\textwidth]{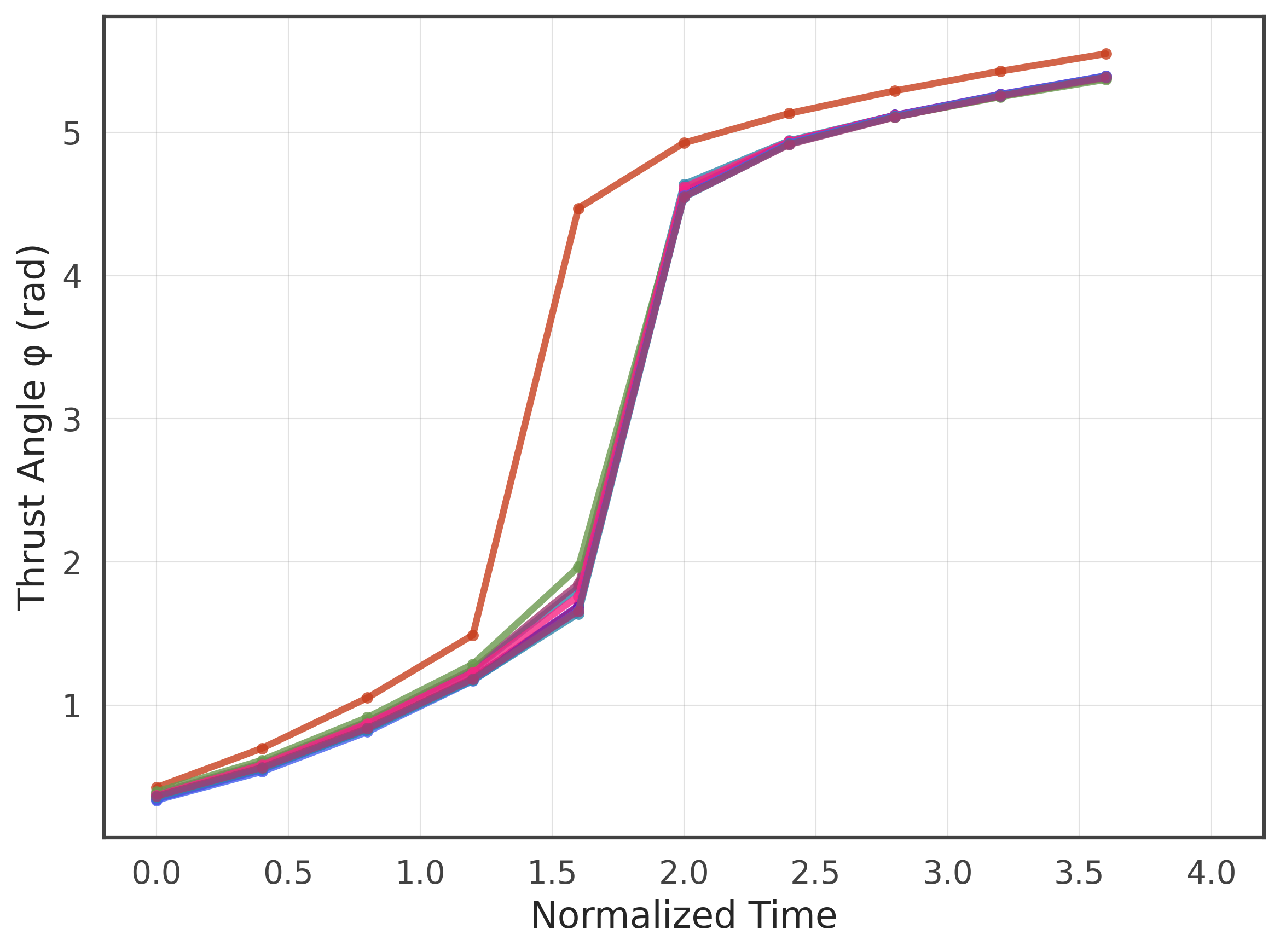}%
  }

  \caption{Orbital transfer trajectories: (a) radial distance evolution, (b) radial velocity component, (c) tangential velocity component, and (d) thrust angle control inputs.}
  \label{fig:orbit_trajectories}
\end{figure*}

With the framework validated on both linear and nonlinear academic control problems, it is now applied to aerospace-relevant scenarios. The orbit raising problem introduces domain-specific challenges absent in the previous systems: nonlinear orbital mechanics governed by Keplerian dynamics, thrust vector control requiring continuous angle optimization, and time-varying mass due to propellant consumption. The controller must manage continuous thrust angle inputs ($\phi \in [0, 2\pi]$ rad) under a fixed thrust magnitude of $T = 0.1405$ N to transfer the spacecraft from an initial orbit to a target radius while respecting the orbital dynamics. Figures~\ref{fig:orbit_grpo} and \ref{fig:orbit_trajectories} present the training and control results.

Figure~\ref{fig:orbit_grpo} displays the GRPO training reward progression for orbital transfer control. The training exhibits distinct phases: initial exploration with rewards near zero (steps 0-200), rapid improvement to rewards around 300 (steps 200-400), and refinement with high variance between steps 400-800. Peak rewards reaching 400-500 demonstrate successful orbital transfers, while occasional drops indicate failed attempts during exploration. The sustained high-reward plateau beyond step 600 with controlled variance suggests convergent policy learning. The higher absolute reward scale compared to previous systems reflects the multi-component reward structure incorporating terminal bonuses, and orbital transfer objectives.

The radial distance trajectories in Figure~\ref{fig:orbit_radius} demonstrate successful orbit raising across multiple scenarios. Starting from an initial radius near $r = 1.0$, all trajectories exhibit smooth, monotonic increases to final radii approaching $r = 1.7-1.9$ over a 4-second maneuver duration. The tight clustering of trajectories indicates consistent transfer performance, with the smooth curvature reflecting efficient energy management under continuous low-thrust propulsion. The absence of oscillations or reversals confirms stable orbital expansion.

Figure~\ref{fig:orbit_radial_vel} shows the radial velocity component evolution during orbital transfer. All trajectories begin near zero radial velocity, increase to peak values of approximately $u \approx 0.45-0.5$ around $t = 2.0$ s, and then decrease back toward zero as the spacecraft approaches the target orbit. This characteristic bell-shaped profile is consistent with optimal low-thrust orbital transfers, where radial velocity must be built up during the transfer and then reduced as the final orbit is approached. The synchronized timing and amplitude of peak velocities across trajectories demonstrate learned coordination between thrust vectoring and orbital mechanics.

The tangential velocity trajectories in Figure~\ref{fig:orbit_tangential_vel} begin at approximately $v \approx 1.0$ and exhibit a characteristic rise-and-fall pattern. Velocities increase to peak values near $v \approx 1.1$ around $t = 1.0-1.5$ s before decreasing to final values of $v \approx 0.65-0.75$. This evolution reflects the coupled nature of orbital dynamics: as the spacecraft transfers to a higher orbit, tangential velocity must decrease to satisfy angular momentum conservation while radial distance increases. The smooth, coordinated modulation of tangential velocity demonstrates the controller's implicit understanding of orbital mechanics constraints.

Figure~\ref{fig:orbit_thrust} presents the thrust angle control sequences, which dictate the direction of applied thrust. Initial thrust angles near $\phi \approx 0.5-1.0$ rad correspond to predominantly tangential thrust to build orbital energy. Around $t = 1.5-2.0$ s, most trajectories exhibit a rapid transition to higher thrust angles ($\phi \approx 5$--$5.5$ rad, approximately $285^{\circ}$--$315^{\circ}$), indicating a strategic shift toward radial or combined thrust vectoring. Some trajectories show intermediate modulation around $\phi \approx 4.5$ rad before settling at final angles. These non-trivial thrust vectoring sequences—substantially different from constant-angle strategies—demonstrate learned optimization of the coupled orbital dynamics, balancing radial expansion with tangential velocity management.

\subsection{Spacecraft Detumbling}

The final and most challenging system in this evaluation combines all prior difficulties, including nonlinear dynamics, multi-dimensional control, and strong coupling between state variables through gyroscopic effects.

\begin{figure}[htb!]
  \centering
  \includegraphics[width=0.9\columnwidth]{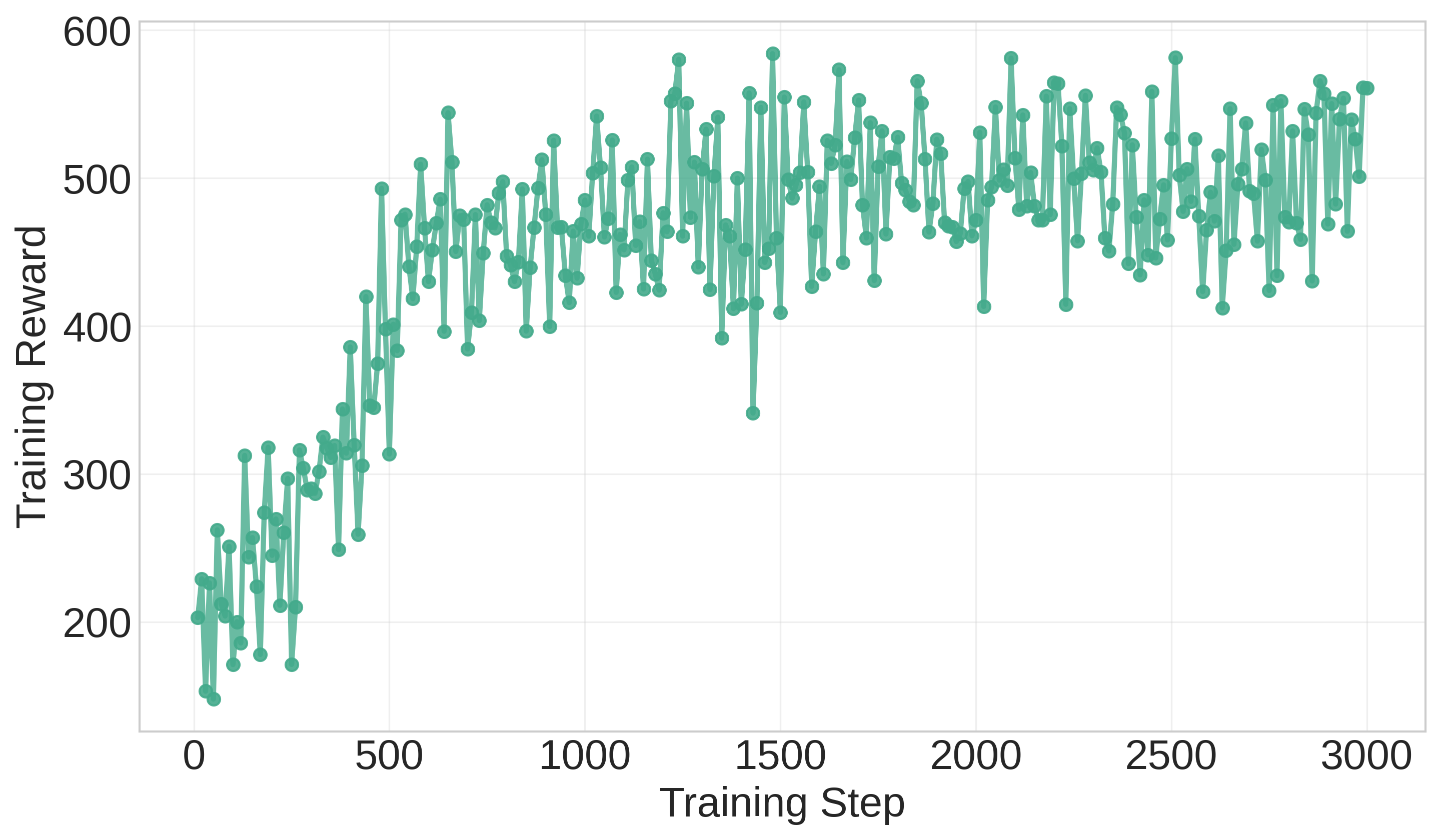}
  \caption{Spacecraft detumbling GRPO training progression showing reward convergence for 3D rotational control with gyroscopic coupling.}
  \label{fig:sd_grpo}
\end{figure}

\begin{figure*}[htb!]
\centering
\subfloat[Angular velocity $\omega_1$\label{fig:sd_omega1}]{%
  \includegraphics[width=0.3\textwidth]{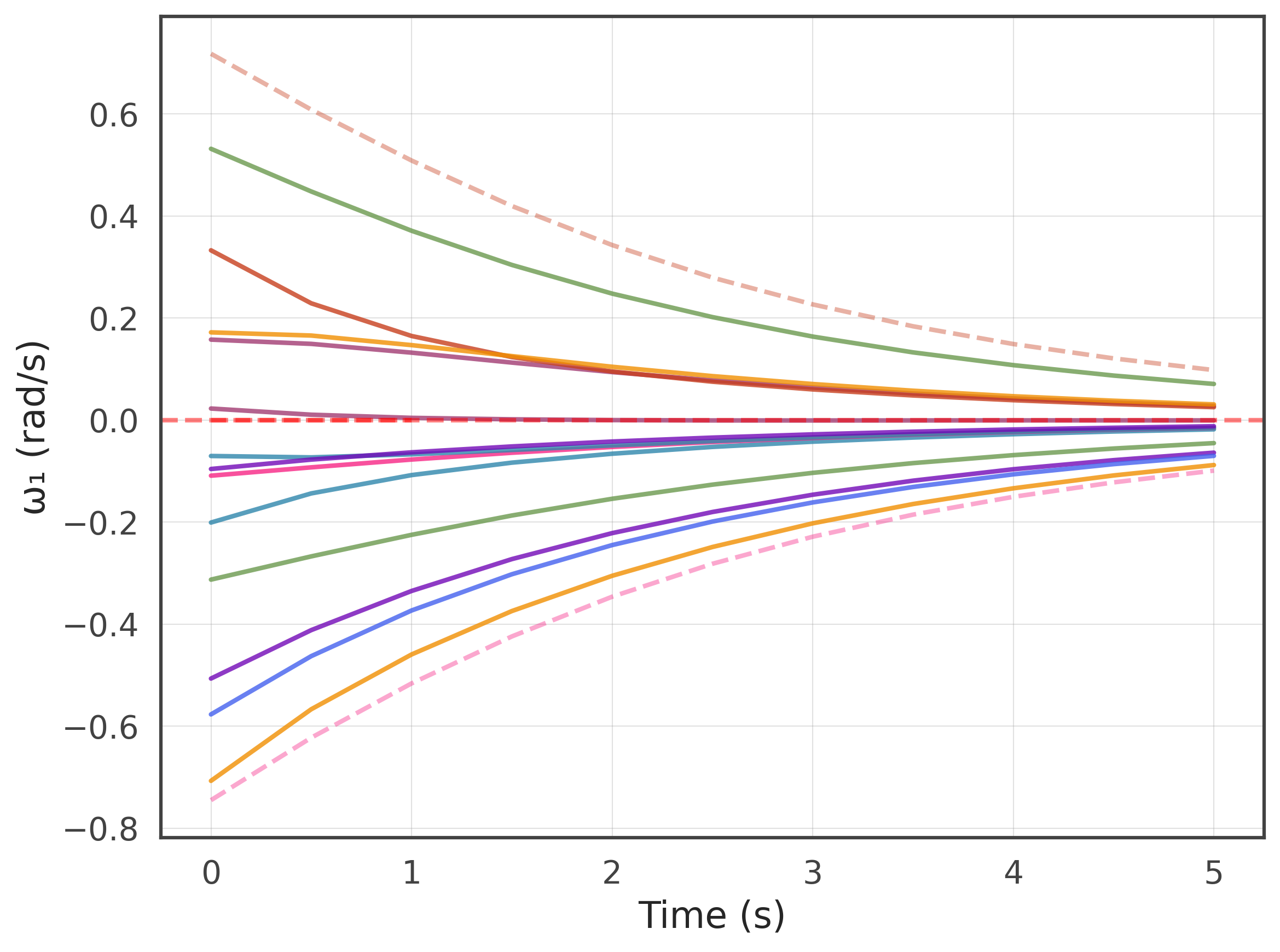}%
}\hspace{0.5cm}%
\subfloat[Angular velocity $\omega_2$\label{fig:sd_omega2}]{%
  \includegraphics[width=0.3\textwidth]{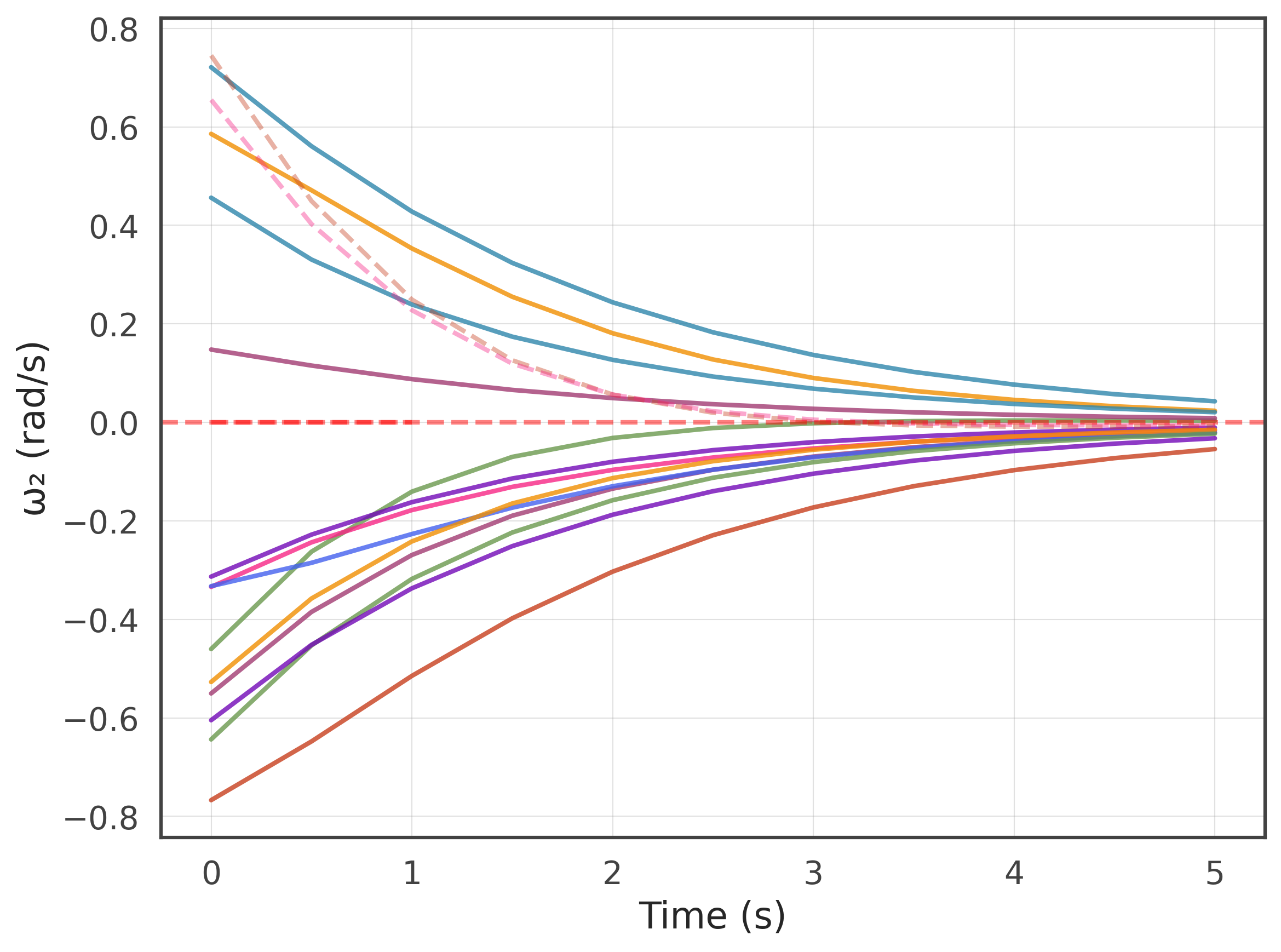}%
}\hspace{0.5cm}%
\subfloat[Angular velocity $\omega_3$\label{fig:sd_omega3}]{%
  \includegraphics[width=0.3\textwidth]{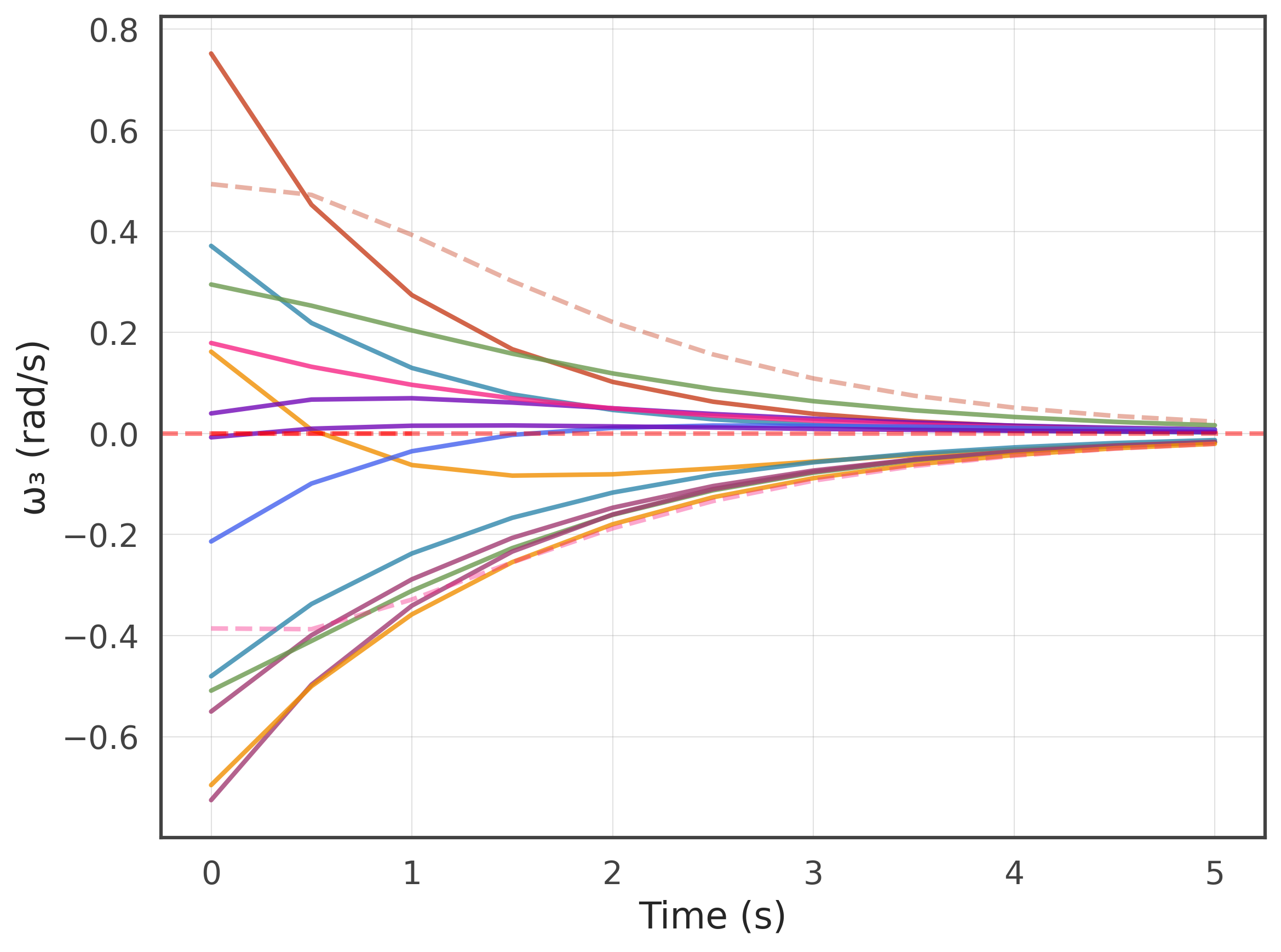}%
}

\vspace{0.3cm}

\subfloat[Control torque $u_1$\label{fig:sd_u1}]{%
  \includegraphics[width=0.3\textwidth]{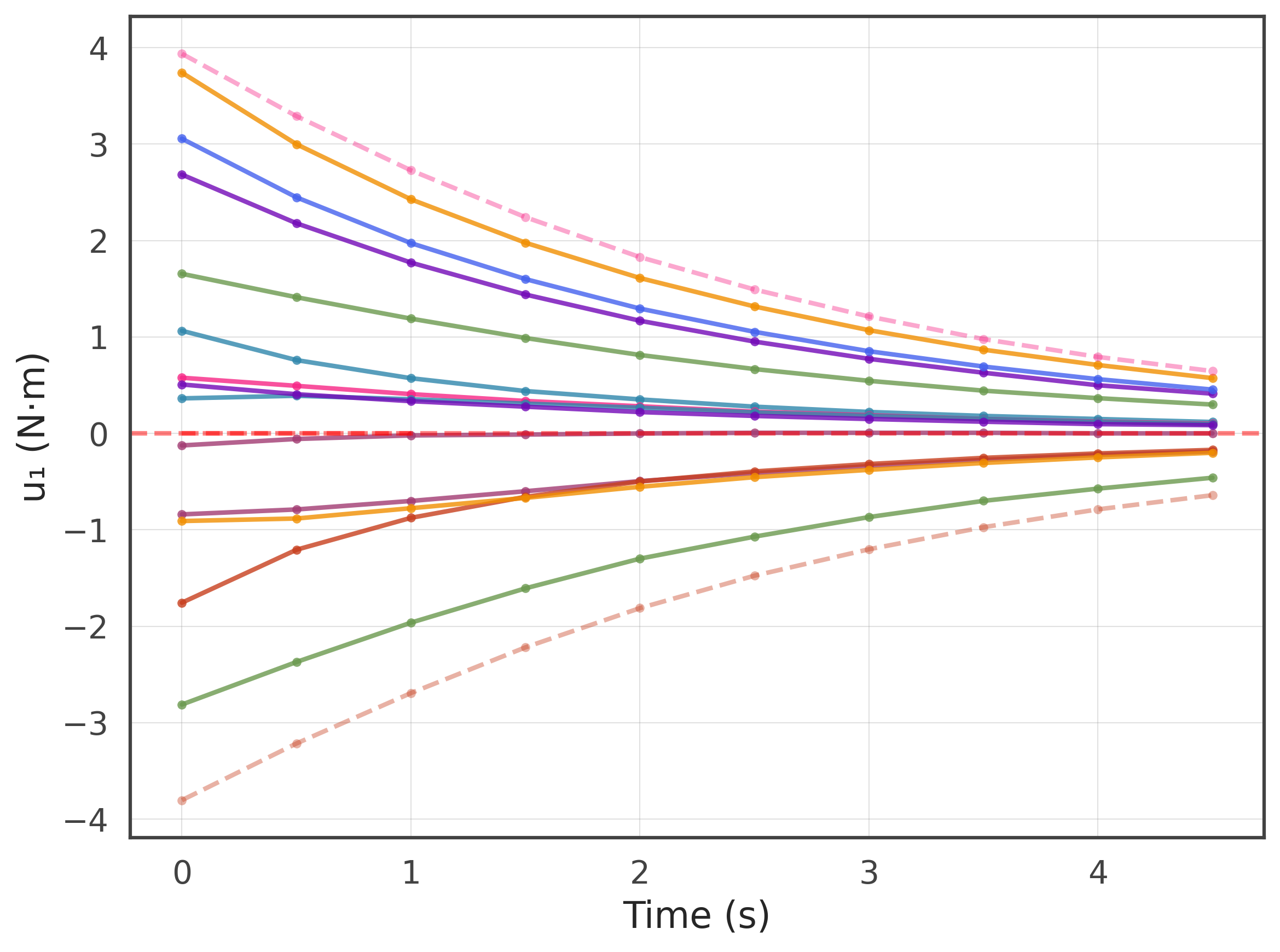}%
}\hspace{0.5cm}%
\subfloat[Control torque $u_2$\label{fig:sd_u2}]{%
  \includegraphics[width=0.3\textwidth]{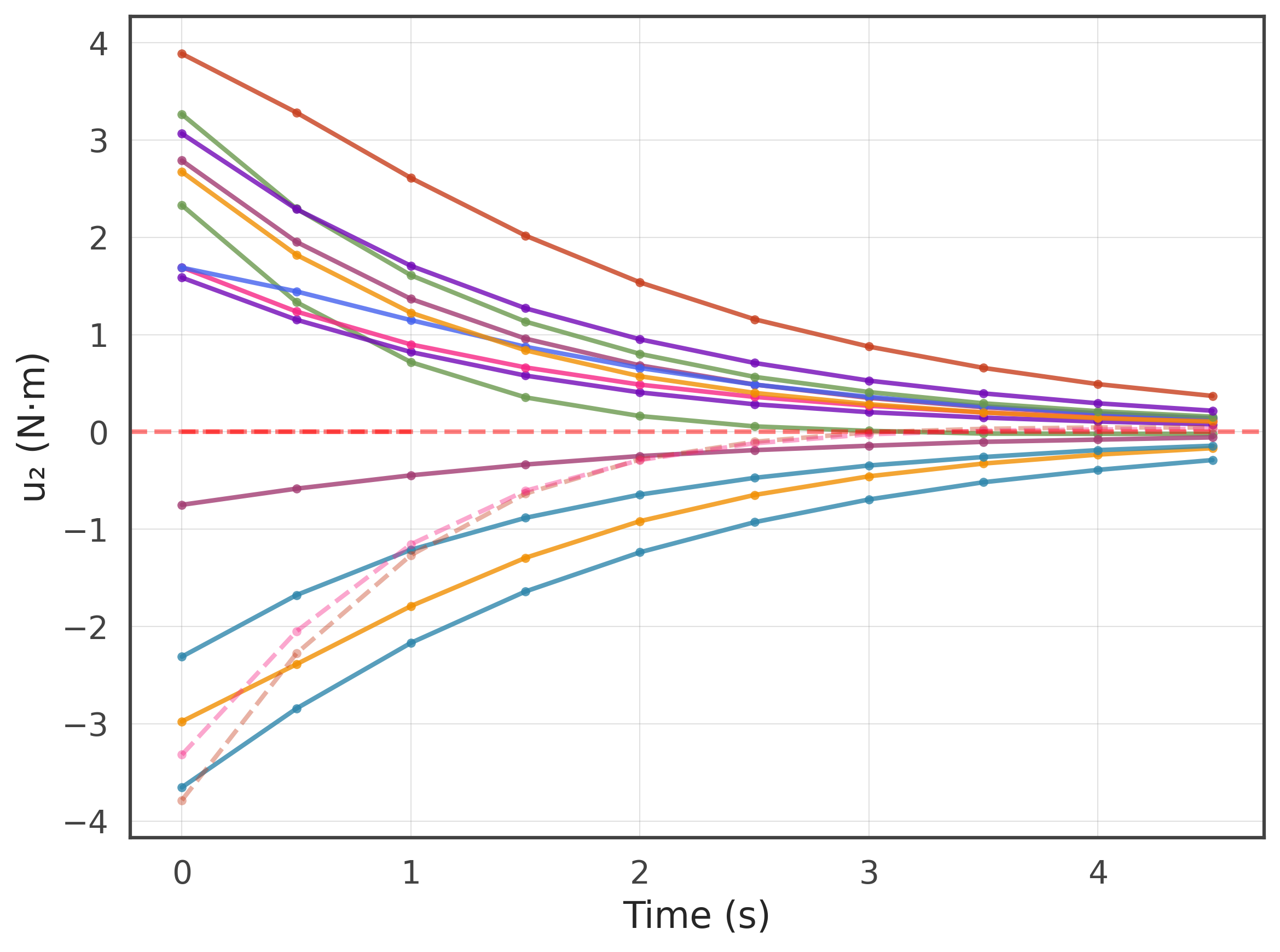}%
}\hspace{0.5cm}%
\subfloat[Control torque $u_3$\label{fig:sd_u3}]{%
  \includegraphics[width=0.3\textwidth]{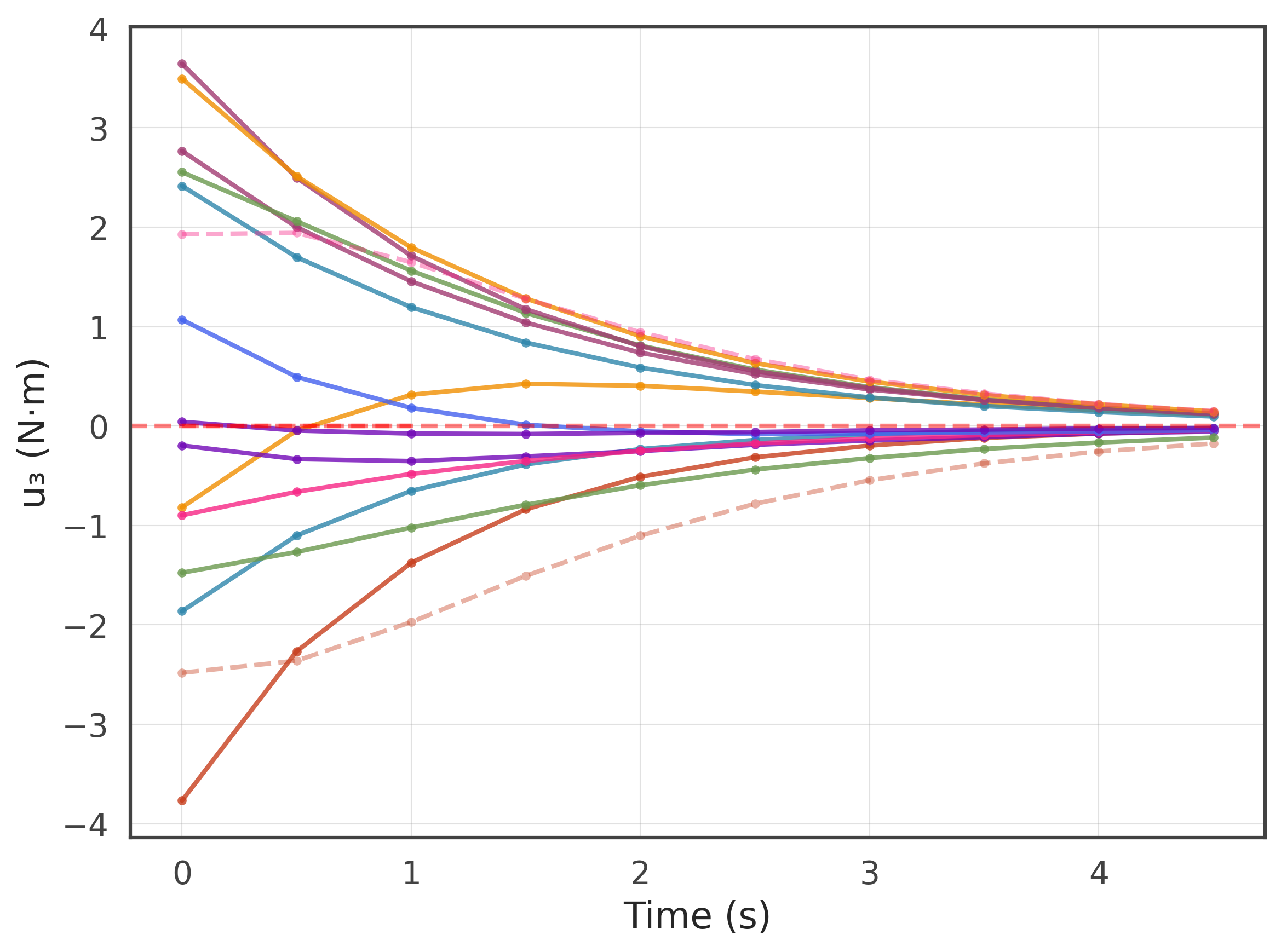}%
}

\caption{Spacecraft detumbling performance: (a-c) three-axis angular velocity trajectories showing coordinated multi-axis stabilization, and (d-f) control torque sequences demonstrating exploitation of gyroscopic coupling effects.}
\label{fig:sd_trajectories}
\end{figure*}

The spacecraft detumbling problem represents the most complex control challenge in this study, requiring simultaneous three-axis rotational stabilization under nonlinear Euler dynamics with gyroscopic coupling. The controller must coordinate three independent torque inputs ($u_1, u_2, u_3 \in [-4, 4]$ N$\cdot$m) to arrest tumbling motion across all axes while respecting the spacecraft's asymmetric inertia tensor $J = \text{diag}([14, 10, 8])$ kg$\cdot$m$^2$. Figures~\ref{fig:sd_grpo} and \ref{fig:sd_trajectories} present the training and control performance.

Figure~\ref{fig:sd_grpo} shows the GRPO training evolution for the detumbling task. Training begins with highly negative rewards around $-60$ to $-80$, indicating initial failure to stabilize the coupled rotational dynamics. Significant improvement occurs between steps 50-150, with rewards climbing above $-40$. The training exhibits substantial variance throughout, with rewards fluctuating between $-80$ and $+20$, and occasional extreme dips below $-80$ even in later stages. Despite this volatility, a general upward trend emerges, with later training steps (400-500) showing improved baseline performance around $-20$ to $0$ and occasional successes reaching $+20$. The persistent high variance reflects the inherent difficulty of learning coordinated multi-axis control under nonlinear gyroscopic coupling, where small control errors can lead to divergent behavior.

The angular velocity trajectories in Figures~\ref{fig:sd_omega1}, \ref{fig:sd_omega2}, and \ref{fig:sd_omega3} demonstrate successful multi-axis stabilization across diverse initial conditions. Figure~\ref{fig:sd_omega1} shows $\omega_1$ trajectories starting from the full range of $[-1.0, 1.0]$ rad/s, all converging smoothly toward zero within approximately 2.5-3.0 seconds. The trajectories exhibit monotonic or near-monotonic convergence with minimal oscillations, indicating well-damped control about the first principal axis.

Figure~\ref{fig:sd_omega2} presents $\omega_2$ evolution from initial conditions spanning $[-1.0, 1.0]$ rad/s. Similar to the first axis, all trajectories converge to near-zero angular velocity within 2.5-3.0 seconds. Several trajectories display slightly more curvature during convergence compared to $\omega_1$, potentially reflecting cross-coupling effects from the intermediate inertia value ($J_2 = 10$ kg$\cdot$m$^2$).

Figure~\ref{fig:sd_omega3} shows stabilization of the third axis from initial rates between $[-1.0, 1.0]$ rad/s. The convergence characteristics are comparable to the other axes, with all trajectories reaching near-zero angular velocity within the same 2.5-3.0 second timeframe. The coordinated simultaneous convergence across all three axes confirms that the learned policy successfully manages the coupled nonlinear dynamics rather than stabilizing axes sequentially.

The control torque trajectories in Figures~\ref{fig:sd_u1}, \ref{fig:sd_u2}, and \ref{fig:sd_u3} reveal sophisticated exploitation of gyroscopic coupling. Figure~\ref{fig:sd_u1} shows $u_1$ control inputs predominantly in the range of $[-2, 2]$ N$\cdot$m, with most trajectories maintaining relatively constant torque values throughout the maneuver rather than smoothly decaying to zero. Several trajectories exhibit near-constant positive torques around $+1.5$ to $+2.0$ N$\cdot$m, while others maintain negative torques near $-0.5$ to $-2.0$ N$\cdot$m. This sustained control effort pattern differs markedly from the decaying controls observed in single-axis or uncoupled systems.

Figure~\ref{fig:sd_u2} displays $u_2$ control trajectories spanning approximately $[-2, 2]$ N$\cdot$m. Similar to $u_1$, many trajectories maintain relatively steady torque values, with distinct clusters around $+1.0$ to $+1.5$ N$\cdot$m and $-0.5$ to $-1.5$ N$\cdot$m. Some trajectories show gradual drift or minor modulation over time, but the predominant behavior is sustained torque application.

Figure~\ref{fig:sd_u3} shows $u_3$ control inputs ranging from approximately $[-1.5, 1.5]$ N$\cdot$m, a slightly narrower range than the other axes. The control patterns exhibit more variability than $u_1$ and $u_2$, with several trajectories showing gradual transitions between different torque levels during the maneuver.

The sustained, non-decaying control torque patterns across all three axes provide strong evidence that the learned policy exploits gyroscopic coupling effects. Rather than applying direct damping torques that would naturally decay as angular velocities decrease, the controller maintains strategic torque combinations that leverage momentum transfer between axes through the products of inertia in Euler's equations. This sophisticated behavior, maintaining seemingly steady control inputs while all three angular velocities simultaneously converge to zero, demonstrates that the LLM-GRPO framework has implicitly learned the underlying physics of coupled rotational dynamics. All control inputs remain well within the $[-4, 4]$ N$\cdot$m constraints while achieving coordinated stabilization.

\subsection{Cross-System Synthesis}

Across all four systems, a consistent pattern emerges: the two-stage training approach successfully establishes control capability through SFT, which GRPO then refines through environmental interaction. The framework scales from two-dimensional linear dynamics to three-dimensional nonlinear spacecraft control without architectural modifications, requiring only system-specific reward tuning and prompt engineering. Convergence times remain comparable (2--4 seconds for stabilization tasks, 4 seconds for orbital transfer), suggesting the learned policies achieve similar relative performance across complexity levels. Notably, the training variance increases with system complexity, from stable convergence in the double integrator to higher variance in spacecraft detumbling, reflecting the inherent difficulty of learning coordinated multi-axis control under nonlinear coupling.

\subsection{Reasoning Output Analysis}
\label{subsec:reasoning_analysis}

A distinguishing feature of this LLM-based control framework is the generation of human-readable reasoning that accompanies each control sequence. Unlike traditional optimal control methods that produce only numerical outputs, this approach provides interpretable explanations of the control strategy, enabling verification and building trust in autonomous systems. Representative reasoning outputs from the trained models are presented below to illustrate the sophistication of learned domain knowledge.

\textbf{Spacecraft Detumbling Example:}
For a spacecraft with initial angular velocities $[\omega_1=-0.507, \omega_2=-0.313, \omega_3=0.040]$ rad/s, the GRPO-trained model generates:

\begin{llmresponse}
<REASONING>
For this spacecraft detumbling maneuver starting with angular velocities
[omega_1=-0.507, omega_2=-0.313, omega_3=0.040] rad/s, I'm using BVP optimal
control to bring the spacecraft to rest in 5.00 seconds.

The spacecraft dynamics follow Euler's rotational equations:
omega_dot = -J^(-1)(omega x J*omega) + J^(-1)*u
With inertia matrix J = diag([14.0, 10.0, 8.0]) kg*m^2.

Analysis:
- Initial angular momentum magnitude: 0.594 rad/s
- Dominant tumbling axis: X (omega_1)
- Coupling constants: K_1=0.143, K_2=-0.600, K_3=0.500

Strategy: optimal
- Apply 3D torque sequence over 10 steps
- Each step duration: 0.5s
- Target: Zero angular velocity (detumbled state)
- Constraints: |omega_i| <= 0.8 rad/s, |u_i| <= 4.0 N*m

This approach exploits the nonlinear coupling between axes while minimizing
control effort and respecting physical constraints.
</REASONING>
<CONTROLS>
[1.973, 1.133, -0.159]
[1.679, 0.897, -0.100]
[1.432, 0.715, -0.058]
[1.228, 0.577, -0.030]
...
</CONTROLS>
\end{llmresponse}

This example achieved a final state error of 0.006, successfully detumbling the spacecraft within the specified time horizon.

\textbf{Orbit Raising Example:}
For orbital transfer from initial state $[r=1.008, u=-0.006, v=0.989]$, the model demonstrates energy-based reasoning:

\begin{llmresponse}
<REASONING>
Energy-Based Orbit Raising Analysis

Current Orbital Energy: epsilon_0 = -0.5034
- Kinetic energy component: (u^2+v^2)/2 = 0.4887
- Potential energy component: -mu/r = -0.9922
- Total specific energy: -0.5034

Target Orbital Energy: epsilon_target = -mu/(2*r_target) = -0.3333
- Required energy increase: Delta_epsilon = 0.1701

Energy Transfer Strategy:
Must add 0.1701 units of specific energy through thrust work

Optimal Thrust Direction:
- Tangential thrust component: Directly increases orbital energy (most efficient)
- Radial thrust component: Minimal, maintain near-circular shape

Physical Constraints:
- Gravitational acceleration at current radius: g = mu/r^2 = 0.9845
- Centrifugal effect: v^2/r = 0.9699
- Net acceleration balance determines thrust requirements

Trajectory Evolution: Over 4.0s, the 10 thrust pulses will incrementally
add energy while maintaining circular shape, progressively raising the orbit.
</REASONING>
<CONTROLS>
0.395, 0.614, 0.912, 1.282, ...
</CONTROLS>
\end{llmresponse}

The reasoning outputs demonstrate several key capabilities learned through the two-stage training process: correct identification of system-specific physics (Euler's equations for rotational dynamics, orbital mechanics for spacecraft maneuvering), quantitative analysis of initial conditions and target states, strategic planning with explicit constraint consideration, and physics-informed justification of control approaches. Notably, the average reasoning length scales with system complexity (approximately 620 characters for the double integrator versus over 1050 characters for orbit raising), suggesting the model allocates proportionally more reasoning effort to physically demanding problems.

Beyond interpretability, GRPO training yields measurable performance improvements. The double integrator achieves average final state error of 0.054, the Van der Pol oscillator achieves 0.039, and spacecraft detumbling achieves 0.034 rad/s. These errors on the order of $10^{-2}$ demonstrate successful convergence across all system types, from linear dynamics to nonlinear multi-axis stabilization under gyroscopic coupling. The results confirm that reinforcement learning refinement through GRPO produces effective control policies while preserving the reasoning capabilities established during supervised fine-tuning.

\section{Conclusion}
\label{sec:conclusion}

This paper introduced a framework combining Large Language Models with Group Relative Policy Optimization to enable reasoning-driven autonomous control. The two-stage methodology, supervised fine-tuning followed by GRPO reinforcement learning, demonstrates that LLMs can generate effective control policies while providing human-readable explanations of their decision-making. Results across four systems of increasing complexity validate this approach: the double integrator achieves rapid state convergence, the Van der Pol oscillator successfully suppresses limit-cycle dynamics, the orbit raising maneuver demonstrates learned thrust vectoring for orbital transfer, and spacecraft detumbling achieves coordinated three-axis stabilization despite nonlinear gyroscopic coupling.

The generated reasoning traces reveal sophisticated understanding of system physics, from limit cycle suppression in nonlinear oscillators to gyroscopic coupling exploitation in spacecraft attitude control. This interpretability represents a significant advancement for safety-critical applications where transparent decision-making is essential.

Future work will prioritize hardware validation on spacecraft attitude control testbeds and investigation of multi-system transfer learning, where a single LLM could control diverse systems without separate fine-tuning. Longer-term directions include integration with model predictive control for real-time replanning and development of formal verification methods for safety guarantees. The framework established here provides a foundation for applying reasoning-capable LLMs to autonomous control, with promising implications for next-generation spacecraft autonomy.

\section*{Acknowledgments}
Research was sponsored by the Department of the Air Force Artificial Intelligence Accelerator and was accomplished under Cooperative Agreement Number FA8750-19-2-1000. The views and conclusions contained in this document are those of the authors and should not be interpreted as representing the official policies, either expressed or implied, of the Department of the Air Force or the U.S. Government. The U.S. Government is authorized to reproduce and distribute reprints for Government purposes notwithstanding any copyright notation herein. 


\bibliographystyle{plainnat}
\bibliography{references}

\begin{thebibliography}{20}
\providecommand{\natexlab}[1]{#1}
\providecommand{\url}[1]{\texttt{#1}}
\expandafter\ifx\csname urlstyle\endcsname\relax
  \providecommand{\doi}[1]{doi: #1}\else
  \providecommand{\doi}{doi: \begingroup \urlstyle{rm}\Url}\fi

\bibitem[{AI Engineering Academy}(2024)]{aiengineering2024grpo}
{AI Engineering Academy}.
\newblock Theory behind grpo.
\newblock \emph{AI Engineering Academy}, 2024.
\newblock URL
  \url{https://aiengineering.academy/LLM/TheoryBehindFinetuning/GRPO/}.

\bibitem[Bai et~al.(2024)]{qwen2024}
Jinze Bai et~al.
\newblock Qwen technical report.
\newblock \emph{arXiv preprint arXiv:2309.16609}, 2024.

\bibitem[Carrasco et~al.(2024)Carrasco, Rodriguez-Fernandez, and
  Linares]{carrasco2024finetuning}
Alejandro Carrasco, Victor Rodriguez-Fernandez, and Richard Linares.
\newblock Fine-tuning llms for autonomous spacecraft control: A case study
  using kerbal space program.
\newblock \emph{arXiv preprint arXiv:2408.08676}, 2024.
\newblock URL \url{https://arxiv.org/abs/2408.08676}.

\bibitem[{DeepSeek AI} and {Guo, Dian and Zhu, Zhihong and Cheng, Wenhao and
  Zhang, Zhaoye and Yuan, Yibin and Li, Lei and Shen, Yichi and Kong, Lv and
  Wen, Yixin and Niu, Qingyan and Jiang, Xin and Zhu,
  Hao}(2025)]{deepseek2025r1}
{DeepSeek AI} and {Guo, Dian and Zhu, Zhihong and Cheng, Wenhao and Zhang,
  Zhaoye and Yuan, Yibin and Li, Lei and Shen, Yichi and Kong, Lv and Wen,
  Yixin and Niu, Qingyan and Jiang, Xin and Zhu, Hao}.
\newblock Deepseek-r1: Incentivizing reasoning capability in llms via
  reinforcement learning.
\newblock \emph{arXiv preprint arXiv:2501.12948}, 2025.
\newblock URL \url{https://arxiv.org/abs/2501.12948}.

\bibitem[{DeepSeek AI} and {Zhu, Zhihong and Guo, Dian and Cheng, Wenhao and
  Fang, Zeming and Yuan, Yibin and Li, Lei and Shen, Yichi and Kong, Lv and
  Wen, Yixin and Zhang, Zhaoye and Jiang, Xin and Niu, Qingyan and Zhu,
  Hao}(2024)]{deepseek2024math}
{DeepSeek AI} and {Zhu, Zhihong and Guo, Dian and Cheng, Wenhao and Fang,
  Zeming and Yuan, Yibin and Li, Lei and Shen, Yichi and Kong, Lv and Wen,
  Yixin and Zhang, Zhaoye and Jiang, Xin and Niu, Qingyan and Zhu, Hao}.
\newblock Deepseekmath: Pushing the limits of mathematical reasoning in open
  language models.
\newblock \emph{arXiv preprint arXiv:2402.03300}, 2024.
\newblock URL \url{https://arxiv.org/abs/2402.03300}.

\bibitem[Gaudet et~al.(2020)Gaudet, Linares, and Furfaro]{gaudet2020adaptive}
Brian Gaudet, Richard Linares, and Roberto Furfaro.
\newblock Adaptive guidance and integrated navigation with reinforcement
  meta-learning.
\newblock \emph{Acta Astronautica}, 169:\penalty0 180--190, 2020.

\bibitem[Hu et~al.(2022)Hu, Shen, Wallis, Allen-Zhu, Li, Wang, Wang, Chen,
  et~al.]{hu2022lora}
Edward~J Hu, Yelong Shen, Phillip Wallis, Zeyuan Allen-Zhu, Yuanzhi Li, Shean
  Wang, Lu~Wang, Weizhu Chen, et~al.
\newblock Lora: Low-rank adaptation of large language models.
\newblock \emph{ICLR}, 1\penalty0 (2):\penalty0 3, 2022.

\bibitem[Jain and Linares(2025)]{jain2025tiny}
Amit Jain and Richard Linares.
\newblock Tiny recursive control: Iterative reasoning for efficient optimal
  control.
\newblock \emph{arXiv preprint arXiv:2512.16824}, 2025.

\bibitem[Jain et~al.(2023)Jain, Eapen, and Singla]{jain2023sparse}
Amit Jain, Roshan Eapen, and Puneet Singla.
\newblock Sparse approximate hamilton-jacobi solutions for optimal feedback
  control with terminal constraints.
\newblock In \emph{2023 62nd IEEE Conference on Decision and Control (CDC)},
  pages 1269--1274. IEEE, 2023.

\bibitem[Jain et~al.(2025)Jain, Rodriguez-Fernandez, and
  Linares]{jain2025multi}
Amit Jain, Victor Rodriguez-Fernandez, and Richard Linares.
\newblock Multi-phase spacecraft trajectory optimization via transformer-based
  reinforcement learning.
\newblock \emph{arXiv preprint arXiv:2511.11402}, 2025.

\bibitem[Kim et~al.(2024)]{kim2024survey}
Yeseung Kim et~al.
\newblock A survey on integration of large language models with intelligent
  robots.
\newblock \emph{arXiv preprint arXiv:2404.09228}, 2024.
\newblock URL \url{https://arxiv.org/abs/2404.09228}.

\bibitem[Kirk(2004)]{kirk2004optimal}
Donald~E Kirk.
\newblock \emph{Optimal control theory: an introduction}.
\newblock Dover Publications, 2004.

\bibitem[Miller and Linares(2019)]{miller2019low}
David Miller and Richard Linares.
\newblock Low-thrust optimal control via reinforcement learning.
\newblock In \emph{29th AAS/AIAA Space Flight Mechanics Meeting}, 2019.

\bibitem[Oestreich et~al.(2021)Oestreich, Linares, and
  Gondhalekar]{oestreich2021autonomous}
Charles Oestreich, Richard Linares, and Ravi Gondhalekar.
\newblock Autonomous six-degree-of-freedom spacecraft docking maneuvers via
  reinforcement learning.
\newblock \emph{Journal of Aerospace Information Systems}, 18\penalty0
  (7):\penalty0 417--428, 2021.

\bibitem[Schulman et~al.(2017)Schulman, Wolski, Dhariwal, Radford, and
  Klimov]{schulman2017proximal}
John Schulman, Filip Wolski, Prafulla Dhariwal, Alec Radford, and Oleg Klimov.
\newblock Proximal policy optimization algorithms.
\newblock \emph{arXiv preprint arXiv:1707.06347}, 2017.

\bibitem[Sutton and Barto(2018)]{sutton2018reinforcement}
Richard~S Sutton and Andrew~G Barto.
\newblock \emph{Reinforcement learning: An introduction}.
\newblock MIT press, 2018.

\bibitem[{Unsloth AI}(2025)]{unsloth2025r1}
{Unsloth AI}.
\newblock Train your own r1 reasoning model locally (grpo).
\newblock Unsloth AI Blog, Jan 2025.
\newblock URL \url{https://unsloth.ai/blog/r1-reasoning}.

\bibitem[Vallado(2001)]{vallado2001fundamentals}
David~A Vallado.
\newblock \emph{Fundamentals of astrodynamics and applications}.
\newblock Springer, 2001.

\bibitem[Wie(2008)]{wie2008space}
Bong Wie.
\newblock \emph{Space vehicle dynamics and control}.
\newblock AIAA, 2008.

\bibitem[Zucchelli et~al.(2025)]{zucchelli2025finetuned}
Enrico Zucchelli et~al.
\newblock Fine-tuned language models as space systems controllers.
\newblock \emph{arXiv preprint arXiv:2501.16588}, 2025.
\newblock URL \url{https://arxiv.org/abs/2501.16588}.

\end{thebibliography}

\end{document}